\documentclass[journal]{IEEEtran}
\usepackage[caption=false,font=normalsize,labelfont=sf,textfont=sf]{subfig}
\usepackage[T1]{fontenc}
\usepackage{graphicx}
\usepackage{times}
\usepackage{helvet}
\usepackage{courier}
\usepackage{amsmath}
\usepackage{algorithm}
\usepackage{algorithmic}
\usepackage{csquotes} 
\usepackage{color}
\usepackage{paralist}
\usepackage{amssymb}
\usepackage{indentfirst}
\usepackage{float}
\usepackage{multirow}
\usepackage{cite}
\usepackage{mathrsfs}

\usepackage{mathrsfs}
\usepackage[colorlinks, linkcolor=red,  anchorcolor=blue, citecolor=blue]{hyperref}
\usepackage{textcomp,booktabs}
\usepackage{amssymb}
\usepackage{pifont}
\usepackage{colortbl}
\definecolor{mygray}{gray}{.9}
\definecolor{mypink}{rgb}{.99,.91,.95}
\definecolor{mycyan}{cmyk}{.3,0,0,0}

\begin{document}
\title{Learning Spatial-Frequency Transformer for Visual Object Tracking} 

\author{Chuanming Tang, Xiao Wang, \emph{Member, IEEE}, Yuanchao Bai, Zhe Wu, Jianlin Zhang, Yongmei Huang
\thanks{Chuanming Tang, Jianlin Zhang and Yongmei Huang are with University of Chinese Academy of Sciences, Beijing, 108408, China, also with Key Laboratory of Optical Engineering, Institute of Optics and Electronics, Chinese Academy of Sciences, Chengdu, 610209, China (e-mail: tangchuanming19@mails.ucas.ac.cn; \{jlin; huangym\}@ioe.ac.cn)}
\thanks{Xiao Wang is with School of Computer Science and Technology, Anhui University, Hefei 230601, China (e-mail: wangxiaocvpr@foxmail.com)}   
\thanks{Yuanchao Bai is with School of Computer Science and Technology, Harbin Institute of Technology, Harbin, 150001, China (e-mail: yuanchao.bai@hit.edu.cn) } 
\thanks{Zhe Wu is with Pengcheng Laboratory, Shenzhen, 518038, China (e-mail: wuzh02@pcl.ac.cn)}

\thanks{Corresponding authors: Xiao Wang; Yongmei Huang}}



\maketitle

\begin{abstract}
Recently, some researchers have begun to adopt the Transformer to combine or replace the widely used ResNet as their new backbone network. As the Transformer captures the long-range relations between pixels well using the self-attention scheme, which complements the issues caused by the limited receptive field of CNN. Although their trackers work well in regular scenarios, they simply flatten the 2D features into a sequence to better match the Transformer. We believe these operations ignore the spatial prior of the target object, which may lead to sub-optimal results only. In addition, many works demonstrate that self-attention is actually a low-pass filter, which is independent of input features or keys/queries. That is to say, it may suppress the high-frequency component of the input features and preserve or even amplify the low-frequency information. 
To handle these issues, in this paper, we propose a unified Spatial-Frequency Transformer that models the Gaussian spatial Prior and High-frequency emphasis Attention (GPHA) simultaneously. To be specific, Gaussian spatial prior is generated using dual Multi-Layer Perceptrons (MLPs) and injected into the similarity matrix produced by multiplying Query and Key features in self-attention. The output will be fed into a softmax layer and then decomposed into two components, i.e., the direct and high-frequency signal. The low- and high-pass branches are rescaled and combined to achieve all-pass, therefore, the high-frequency features will be protected well in stacked self-attention layers.  
We further integrate the Spatial-Frequency Transformer into the Siamese tracking framework and propose a novel tracking algorithm termed SFTransT. The cross-scale fusion based SwinTransformer is adopted as the backbone, and also a multi-head cross-attention module is used to boost the interaction between search and template features. The output will be fed into the tracking head for target localization. 
Extensive experiments on short-term and long-term tracking benchmarks all demonstrate the effectiveness of our proposed framework. 
Source code will be released at \url{https://github.com/Tchuanm/SFTransT.git}.

\end{abstract}

\begin{IEEEkeywords}
Visual Tracking; Gaussian-Prior; Frequency-Emphasis; Spatial-Frequency Transformer; Siamese Network
\end{IEEEkeywords}

\section{Introduction}

\IEEEPARstart{V}{isual} Object Tracking (VOT) is one of the most important computer vision tasks, which targets predicting the position of the target object in the continuous video frames based on initialized bounding box in the first frame. It has been widely used in many practical applications, such as unmanned aerial vehicle tracking~\cite{cao2021feature, lin2020learning}, multi-modal tracking~\cite{li2018learning, ramesh2020tld}, military field, etc. With the help of deep learning techniques~\cite{lecun2015deep, zhu2022eventsnn}, many end-to-end visual trackers are proposed, such as the multi-domain based tracker~\cite{mdnet} and Siamese matching-based tracking~\cite{siamrpn++, siamfc, siamban}. Although these trackers achieve high performance on existing benchmark datasets, their performance is still limited when facing challenging factors like dramatic deformation, heavy or full occlusion, fast motion, etc.

To address the aforementioned issues, current researchers attempt to design powerful neural networks for robust feature representation learning. Early works like SiamFC~\cite{siamfc} and SiamRPN~\cite{siamrpn} pave new ways for Siamese matching-based tracking. The key step is calculating the similarity between the template and the search branch. Usually, ResNet\cite{resnet} is adopted as the backbone for feature extraction and correlation computation. More and more Siamese trackers~\cite{siamrpn, siamban, siamcar} are proposed under this framework, but their tracking performance is still poor in challenging scenarios due to the utilization of convolutional layers. Many works demonstrate that the limited receptive field of convolution kernels may only bring sub-optimal results~\cite{vit, swin}.

\begin{figure}[t]
\centering
\includegraphics[width=1\columnwidth]{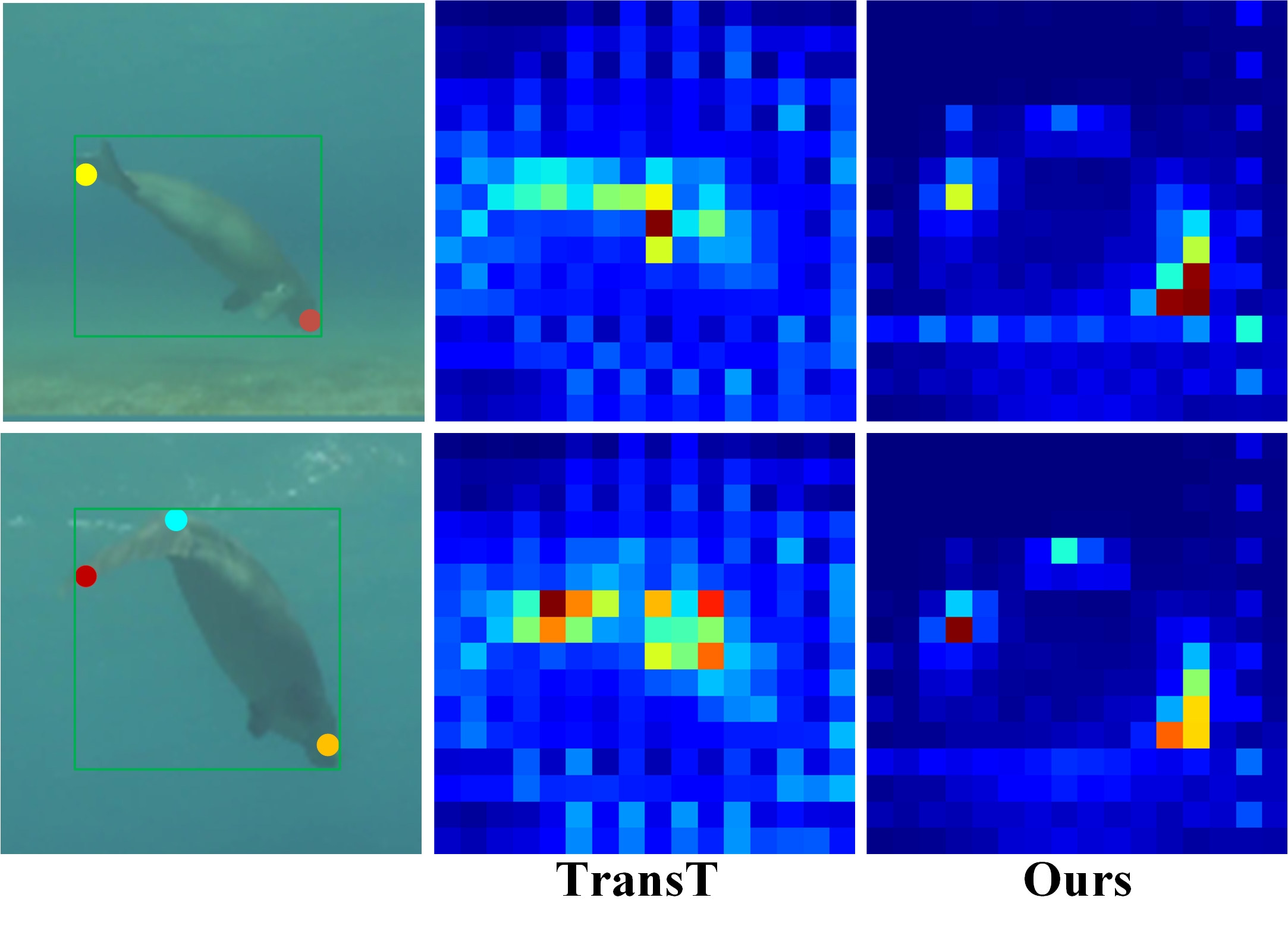} 
\caption{ Comparison of the activation maps estimated by our proposed Spatial-Frequency Transformer and existing state-of-the-art Transformer-based visual tracker TransT~\cite{transt}. Note that, our model pays more attention to the edge position and specific target components (e.g. head and tail) rather than the less discriminative structures (e.g. trunk).} 
\label{attnmap_comparision}
\end{figure}

Recently, the Transformer is becoming a hot research topic in both Natural Language Processing (NLP)~\cite{attentionisallyouneed} and Computer Vision (CV)~\cite{vit}. Its core module is termed self-attention, which models long-range relations well. Some researchers have already exploited the combination of Transformer and CNN networks, such as TransT~\cite{transt}, TrDIMP~\cite{trdimp}, STARK~\cite{stark}, and obtained better results than full CNN-based trackers. These works mainly focus on fusing the template and search regions using Transformer layers. 
The original Transformer mechanism simply flattens the two-dimensional feature maps into one-dimensional feature sequences and calculates the attention scores between each patch. 
\textcolor{black}{
However, such a 2D-to-1D transformation destroys the spatial distribution of the target object and loses the spatial dependency information of each patch.
} 
Therefore, how to retain the spatial prior of the target object when learning the long-range relations based on Transformer is an urgent problem to be solved. 

On the other hand, more and more works~\cite{cai2020note, wang2022anti} demonstrate that the self-attention mechanism is actually a low-pass filter, which is independent of input features or keys/queries. In other words, the Transformer may suppress the high-frequency component of input features and preserve or even amplifies the low-frequency information. \textcolor{black}{ In the stacked process of Transformer layers, high-frequency information will drop layer-by-layer and resulting in over-smoothing features. As the layer number increases infinitely, the final output will dilute high-frequency components and only keep the DC bias. When the deep attention layer passes the DC bias merely, this attention block will degenerate into a linear projection layer, making the model fall into performance saturation and loses feature expressive power. In this saturation, with deformation happening, the original Transformer deep layers will lack the ability to find high-frequency commonality of the deformed target with the template. When facing distractors, it will also lose the discriminate ability of high-frequency feature differences between target and distractors, further resulting in target loss and tracking failure in the tracking process.
}
\textcolor{black}{At the same time, current Transformer based trackers focus on framework structure, augmentation module, and tracking strategy but employ the original attention blocks directly. In this situation, targeting the above defect of spatial and frequency aspects of Transformer mechanism,} it is natural to raise the following question: \emph{how can we complement the spatial prior and the high-frequency component of the target sequence when using the deeply stacked multi-head self-attention modules?}

\begin{figure*}
\centering
\includegraphics[width=1\textwidth]{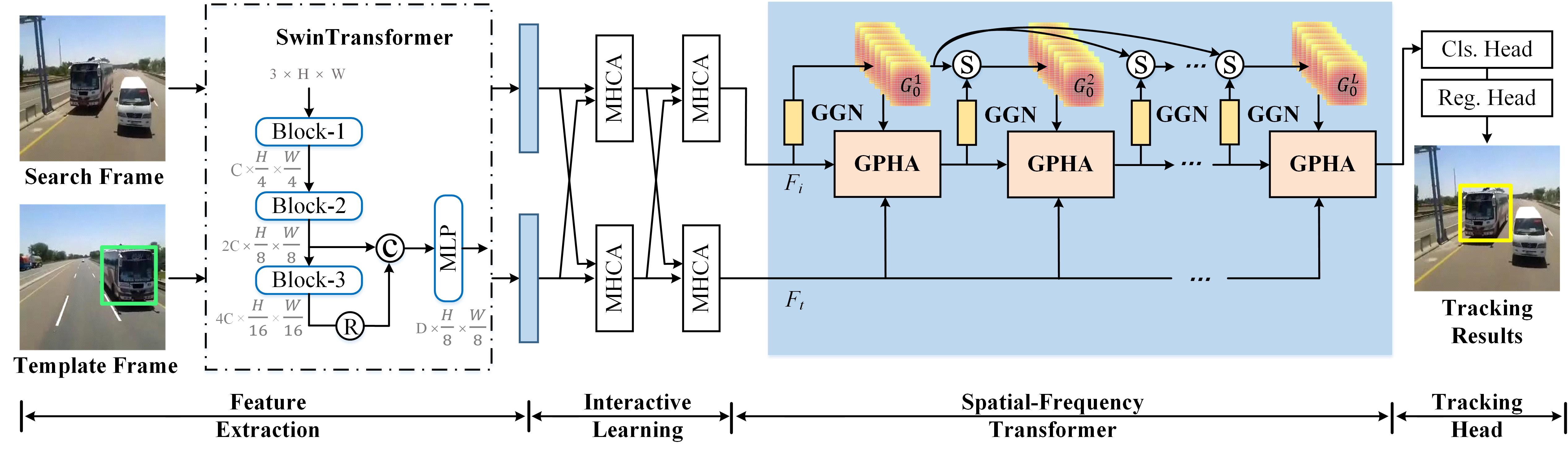} 
\caption{An overview of our proposed SFTransT visual object tracking algorithm. It follows the Siamese matching framework which takes the template and search frame as input. The Swin-Tiny network is adopted as the backbone, and the cross-scale features are fused as embedded features. Then, a Multi-Head Cross-Attention (MHCA) module is used to boost the interactions between the dual features. The output will be fed into our core component Spatial-Frequency Transformer, which models the Gaussian spatial prior and low-/high-frequency feature information simultaneously. 
More in detail, the GGN is adopted to predict the Gaussian spatial attention which will be added to the self-attention matrix. Then, the GPHA is designed to decompose them into low- and high-pass branches to achieve all-pass information propagation. 
Finally, the enhanced features will be fed into the classification and regression head for target object tracking. \textcircled{\scriptsize C}, \textcircled{\scriptsize R}, \textcircled{\scriptsize S} represent the Concatenate, Rearrange, and Shift-and-Sum operators, respectively. }  
\label{framework}
\end{figure*}

With this question in mind, in this paper, we design a novel spatial-frequency Transformer framework for visual object tracking, termed \textbf{SFTransT}. Generally speaking, our proposed SFTransT follows the Siamese tracking framework, and it contains four main parts, including the cross-scale fusion-based SwinTransformer, Multi-Head Cross-Attention (MHCA) module, Spatial-Frequency Transformer (SFFormer), and tracking head. 
\textcolor{black}{
Specifically, the cross-scale fusion-based SwinTransformer is used as the backbone network to extract the feature representations of template and search regions. Unlike existing Transformer trackers that adopt the last stage feature and CNN trackers that adopt the layer-wise aggregation strategy, our framework fuses the cross-scale representation of different Transformer layers with a lossless way to keep the multi-stage information of extracted features. 
}
Then, we feed the backbone features into the MHCA module for interactive learning of template and search features.
\textcolor{black}{
The output will be fed into our newly proposed unified spatial-frequency Transformer, which introduces the Gaussian-prior and frequency-emphasis at the same time. Gaussian-prior knowledge keeps the spatial relationship between image patches, and the multi-head Gaussian generation and bias mechanism make the gaussian map adaptively fit the real scenes of each layer. High-frequency emphasis attention can divide the target feature into low and high-frequency components, enlarging the high-frequency component by learnable parameters, further avoiding the high-frequency information dilution of deep Transformer layers. Integrating the spatial-prior attention with high-frequency emphasis attention into the proposed spatial-frequency attention with cascade framework, the proposed Transformer can simultaneously model spatial-frequency without feature saturation.
}
%
Finally, the tracking head is adopted to directly transform the enhanced features into target localization without the widely used correlation operators. An overview of our proposed SFTransT is illustrated in Fig.~\ref{framework}.

To sum up, we  draw the contributions of this paper in the following three aspects:
\begin{itemize}
\item We analyze the limitations of current Transformer based trackers and propose a novel spatial-frequency Transformer for visual object tracking. It simultaneously models the spatial prior and frequency information for more robust feature representation learning. 

\item We propose a cross-scale fusion-based SwinTransformer as the shared backbone network for tracking, which makes our model become a fully Transformer-based visual tracking framework. 

\item Extensive experiments are conducted on nine large-scale and popular tracking benchmark datasets, including LaSOT, TNL2K, TrackingNet, GOT-10K, WebUAV-3M, etc. These experimental results fully validate the effectiveness of our proposed module for tracking, and also demonstrate that our tracker achieves state-of-the-art performance. For example, it obtains an SR of 69.1\% on LaSOT\cite{lasot} and 72.7\% on GOT-10K\cite{got10k}. 
\end{itemize}

\section{Related Work}
In this section, we will give a brief review of the Siamese-based trackers, Transformer based trackers, and Gaussian prior utilized in deep learning. 
More related works can be found in the following survey papers~\cite{survey1, survey3} and paper list~\footnote{\url{https://github.com/wangxiao5791509/Single_Object_Tracking_Paper_List}}. 

\noindent
\textbf{Siamese based Tracking. } 
With the release of SiamFC~\cite{siamfc} which formulated the visual tracking as a similarity matching problem between the template and search region, many subsequent trackers have been developed to gain performance improvement. SiamRPN~\cite{siamrpn} introduced RPN~\cite{faster-rcnn} structure to predict object scale with the variation of aspect ratio. SiamRPN++~\cite{siamrpn++} utilized the deeper network ResNet-50~\cite{resnet} as the parameter-shared backbone and took the representation ability of the tracking framework to new levels. STMtrack~\cite{stmtrack} followed the Siamese framework but did not share the backbone of the template and search branch at all. Wang et al.~\cite{wang2018sintpp} proposed the adversarial positive instance generation for visual tracking. With the object detection and segmentation development, some trackers~\cite{siamban, siamcar, siamgat, ocean, siamfc++} introduce the anchor-free prediction head to cut down the heuristic parameters. SiamR-CNN~\cite{siamr-cnn} joined Faster R-CNN~\cite{faster-rcnn} with Siamese framework, realizing satisfying robustness in scale change and potential distractor object. Pi et al.~\cite{pi2021instance} introduced an instance-based feature pyramid to achieve adaptive feature fusion. Zhang et al.~\cite{zhang2022target} designed a target-distractor-aware model with discriminative enhancement learning loss to learn target representation.

The above Siamese trackers have achieved gratifying performance in attributions of motion blur, deformation, etc., but are still easily affected by similar distractors and heavy occlusion which suppress the robustness of trackers. To solve these issues, some trackers introduced online learning approaches. Martin et al.~\cite{atom, dimp, prdimp} collected the historical hard negative samples to accelerate the discriminate feature learning process and verified the effectiveness of discriminate online tracking. The authors of KeepTrack~\cite{keeptrack} introduced a learned association network to discriminate distractors for continuing robustness tracking. UpdateNet~\cite{updatenet} broke the linear template update strategies and proposed the update network to learn the nonlinear fusion template. TANet~\cite{tanet} adopted a joint local-global search scheme for robust tracking which works well in long-term videos. 
Inspired by these trackers, our proposed SFTransT also follows the Siamese tracking framework. Nevertheless, our model adopts cross-scale fusion-based SwinTransformer~\cite{swin} as the backbone and obtains more robust features for visual object tracking.

\noindent
\textbf{Transformer based Tracking. } 
Vision Transformer (ViT)~\cite{vit} first introduced Transformer~\cite{attentionisallyouneed} into image classification and achieved competitive performance with CNNs. It divided the input images into local patches and flattened feature maps into patch sequences. Motivated by the development of vision Transformers, traditional tracking approaches combined CNN networks with Transformer components. TransT~\cite{transt} first introduced the Transformer structure into the tracking field and utilized the self-attention and cross-attention of DETR~\cite{detr} to enhance the ego-context and cross-feature. TrDiMP~\cite{trdimp} considered the encoder and decoder network into parallel branches to explore the temporal contexts across the two branches. STARK~\cite{stark} followed the encoder-decoder framework and combined the template with search branch into one branch after feature extraction. In this way, STARK removed the cross-fusion operations in Siamese trackers and kept the tracking framework similar to detection. DTT~\cite{dtt} similarly kept encoder-decoder Transformer architecture and dense prediction to build a discriminate tracker. ToMP~\cite{tomp} proposed a Transformer-based model prediction module to replace the traditional optimization-based predictor and capture global relations between frames. Cross-modality Transformer (CMT) was proposed in work~\cite{wang2021visevent} to achieve RGB-Event fusion-based tracking. 
Nie et al.~\cite{nie2022spreading} generate the fine-grained prior knowledge with explicit representation for template target. 

However, these Transformer based trackers are all limited by the context-representation defect of CNN backbone~\cite{alexnet, resnet}. With the release of Swin-Tranformer~\cite{swin}, SwinTrack~\cite{swintrack} evolved the backbone from CNN~\cite{resnet} into Transformer to learn much superior feature extraction and representation capability. CSWinTT~\cite{cswintt} further evolved the attention from pixel-to-pixel to cyclic shifting window level. Moreover, SBT~\cite{sbt} proposed the single branch network independent of the Siamese framework, allowing the template and search features to interact deeply at each stage of feature extraction. MixFormer~\cite{mixformer} further proposed the one-stage one-branch structure and utilized the  CVT~\cite{cvt} for feature extraction and information integration. 
Although existing Transformer driven trackers achieve impressive performance, they may obtain sub-optimal results only due to the ignorance of spatial prior and low-/high-frequency feature learning. In contrast, our proposed SFTransT tracker considers the two aspects in a unified tracking framework, which bring us a higher performance on multiple benchmark datasets.

\textcolor{black}{
At the same time, the multi-scale feature fusion strategy, which is popularly employed in Siamese tracking frameworks, is ignored by Transformer trackers. SiamRPN++~\cite{siamrpn++} adopts the depthwise correlation to fuse the multi-scale feature of ResNet50. Similarly, SiamBAN~\cite{siamban}, SiamCAR~\cite{siamcar}, and SiamFC++~\cite{siamfc++} follow the cross-correlation to gain multiple knowledge representations. While modern Transformer trackers~\cite{transt, swintrack, stark, mixformer, cswintt} focus on powerful fitting ability and merely adopt the last stage features for augmentation and prediction.  However, as Yang et al.~\cite{yang2021multiple} pointed out, multiple knowledge is essential and meaningful for feature extraction, fusion, and representations.  
For feature fusion, TransT~\cite{transt} replaces the classical cross-correlation with a cross-attention module to fuse the last stage features of dual-branch.  SwinTrack~\cite{swintrack} and STARK~\cite{stark} concatenate the dual-branch feature directly. Wang et al.~\cite{wang2020symbiotic} introduce symbiotic attention to encourage interaction between two branches. In our work, we propose the cross-scale feature fusion method for backbone features to represent cross-source and diversified features. 
}

\noindent
\textbf{Gaussian Prior in Deep Learning. }  
Gaussian distribution, also called normal distribution, is widely used in machine learning. Most machine learning models are based on the assumption that data obey Gaussian distribution, such as naive Bayesian classification and GMM~\cite{gmm}. That is to say, the probability of being located near the center is always higher than the position probability of edge position. For natural language generation, D2GPo~\cite{li2020data} proposed to add the 1D data-dependent Gaussian prior objective into maximum likelihood estimation of context, considering the Gaussian prior distribution as the similarity between prediction words. Guo et al.~\cite{guo2019gaussian} introduced 1D Gaussian prior to the self-attention mechanism for language inference, modelling the local structure of sentences. Additionally, it proves that the 1D Gaussian prior probability can correct the distance to the central word.

For the computer vision tasks, NWD~\cite{wang2021normalized} utilized the spatial characteristic and modelled the 2D Gaussian distribution as the bounding box prior knowledge to compute the feature similarity by their corresponding Gaussian distributions. With similarity motivation, Yang et al.~\cite{yang2021rethinking} converted the rotated bounding box into 2D Gaussian distribution and proposed the Gaussian Wasserstein Distance (GWD) to approximate the rotational bounding box. The above detection and language generate works consider the Gaussian prior as the probability density function distribution. Unlike the aforementioned language and detection works, DRAW~\cite{gregor2015draw} applied a Gaussian filter to yield smoothly varying image patches for image generation. All these works inspired us to design novel Gaussian prior for robust visual object tracking.

\section{Methodology}
In this section, we will first give an overview of our proposed tracker in Section~\ref{Overview}. Then, we delve into the details of our spatial-frequency Transformer tracking framework, which contains four main modules: feature extraction module, interactive learning module, spatial-frequency Transformer module, and tracking head. After that, we will talk about the details of training and testing of our tracker in Section~\ref{train_test_phase}.

\subsection{Overview} \label{Overview} 
In this work, we propose a novel visual tracker based on Siamese matching framework, termed SFTransT, as shown in Fig.~\ref{framework}. Given the template and search frame, we design a powerful backbone network for the feature extraction based on cross-scale fusion-based SwinTransformer~\cite{swin}. After obtaining the backbone features, we introduce MHCA (Multi-Head Cross-Attention) modules to boost the interactive learning between the search and template features. Then, the enhanced features will be fed into the spatial-frequency Transformer which considers the Gaussian spatial prior and low-/high-frequency feature information simultaneously. Finally, we utilize a three-layer MLP head to predict the target object position and classification confidence scores. Our network can be trained in an end-to-end manner by jointly optimizing the BCE, CIOU, and L1 loss. In the following subsections, we will give a detailed introduction to these modules respectively.

\subsection{Feature Extraction using Cross-Scale Fusion based SwinTransformer} \label{backbonenetwork}

Given the first video frame $T$ and the $i^{th}$ search images \textit{$I_i$}, we first resize them into $\textit{T} \in \mathbb{R}^{3\times H_T\times W_T}$ and $\textit{$I_i$} \in \mathbb{R}^{3\times H_I\times W_I}$ respectively. Then, these images or patches are fed into a shared backbone network for feature embedding, such as the widely used ResNet~\cite{resnet}. Although good performance can be obtained, more and more works demonstrate that the feature representation may not be optimal due to the limited receptive field caused by local convolutional filters. Inspired by the success of Swin Transformer~\cite{swin} in the image classification task, in this work, we adopt the Swin-Tiny as our backbone network because it achieves a good balance between performance and efficiency. More in detail, the fourth block of Swin-Tiny is removed to gain the effective strides as $8\times$ and reduce parameters from 28.8M to 12.6M. The shift window size is set as 8 to accommodate our tracking task.

To fully utilize the hierarchical features of the Swin-Tiny network, we propose a novel Cross-Scale Fusion (CSF) module to adaptively fuse them for a more robust representation. As shown in Fig.~\ref{framework}, when we input an image whose dimension is $3\times H \times W$ into the Swin-Tiny network, the dimension of features output from the second and third blocks are $F_\textit{2} \in \mathbb{R}^{2C\times\frac{H}{8}\times\frac{W}{8}}$ and $F_\textit{3} \in \mathbb{R}^{4C\times\frac{H}{16}\times\frac{W}{16}}$ respectively. Therefore, the fused features can be obtained via: 
\begin{equation}
F_{CSF} = MLP[F_\textit{2}, Rearrange(F_\textit{3})], 
\end{equation}
where [ , ] denotes the concatenate operation along the channel dimension. $Rearrange$ is used to increase the resolution of feature maps without scale information missing. $MLP$ is consisted of two fully connected (FC) layers to project the feature dimension from 3C to D. 
Therefore, we can get the search and template feature $F_I \in \mathbb{R}^{D\times\frac{H_I}{8}\times\frac{W_I}{8}}$ and $F_T \in \mathbb{R}^{D\times\frac{H_T}{8}\times\frac{W_T}{8}}$. In our experiments, C and D are set as 96 and 256, respectively.

\subsection{Interactive Feature Learning using MHCA}   
After we obtained the backbone features $F_I \in \mathbb{R}^{D\times\frac{H_I}{8}\times\frac{W_I}{8}}$, $F_T \in \mathbb{R}^{D\times\frac{H_T}{8}\times\frac{W_T}{8}}$ for the search and template branch, we feed them into the Multi-Head Cross-Attention (MHCA) module to boost the interactive feature learning. Specifically, the feature maps are firstly reshaped into patch sequences, $ F_I \in \mathbb{R}^{D\times S_I}$ and $ F_T \in \mathbb{R}^{D\times S_T}$, where the $S_I=\frac{H_I\times W_I}{8\times8}$, $S_T=\frac{H_T\times W_T}{8\times8}$. 
Then, the input template features are transformed into query $\mathcal{Q}_T$, key $\mathcal{K}_T$, and values $\mathcal{V}_T$ using non-shared fully connected (FC) layers, i.e., $\mathcal{Q}_T = FC_I(F_{T}), \mathcal{K}_I = FC_I(F_I), \mathcal{V}_I = FC_I(F_I)$. The cross-attention from template to search feature used in $n^{th}$ head $H_I^n$ can be mathematically formulated as:
\begin{flalign}
& H_I^n = CrossAttn^n_{T \rightarrow I}(\mathcal{Q}_T, \mathcal{K}_I, \mathcal{V}_I) \\ 
&     ~~~~= LN(Softmax(\frac{\mathcal{Q}_T  \mathcal{K}_I^T}{\sqrt{c}})\mathcal{V}_I), \\ 
& \hat{F_I} = FFN_I[H_I^1, \ldots, H_I^N],   
\end{flalign}
where $FFN$ denotes the feed-forward network, which targets to magnify the fitting ability of cross-attention. $LN$ is short for layer normalization. $N$ denotes the number of heads used in cross-attention.

Similarly, the cross-attention from search to template features can be obtained by: 
\begin{flalign}
& H_T^n = CrossAttn^n_{I \rightarrow T}(\mathcal{Q}_I, \mathcal{K}_T, \mathcal{V}_T) \\ 
&     ~~~~= LN(Softmax(\frac{\mathcal{Q}_I  \mathcal{K}_T^T}{\sqrt{c}})\mathcal{V}_T), \\ 
& \hat{F_T} = FFN_T[H_T^1, \ldots, H_T^N]. 
\end{flalign}
It is worth noting that the MHCA mentioned above or other Transformer modules all adopt the sinusoidal position embedding~\cite{detr}, which is added to the input sequences.

\subsection{Spatial-Frequency Transformer}

Once we get the output from the MHCA module, we can directly feed them into the tracking head for target localization like many trackers do~\cite{transt, trdimp, dtt}. However, we can find that the aforementioned feature interaction between search and template features is based on 1D feature vectors to make full use of the Transformer layers. Although good performance can already be obtained, the feature enhancement based on feature vectors cannot capture the spatial prior which is also very important for visual object tracking. In addition, many works demonstrate that the Transformer layer is actually a low-frequency passer and may suppress the high-frequency features. To address these issues, we propose a novel Spatial-Frequency Transformer (SFFormer) which will be introduced in the following subsection.

\begin{figure}
\centering
\includegraphics[width=0.5\textwidth]{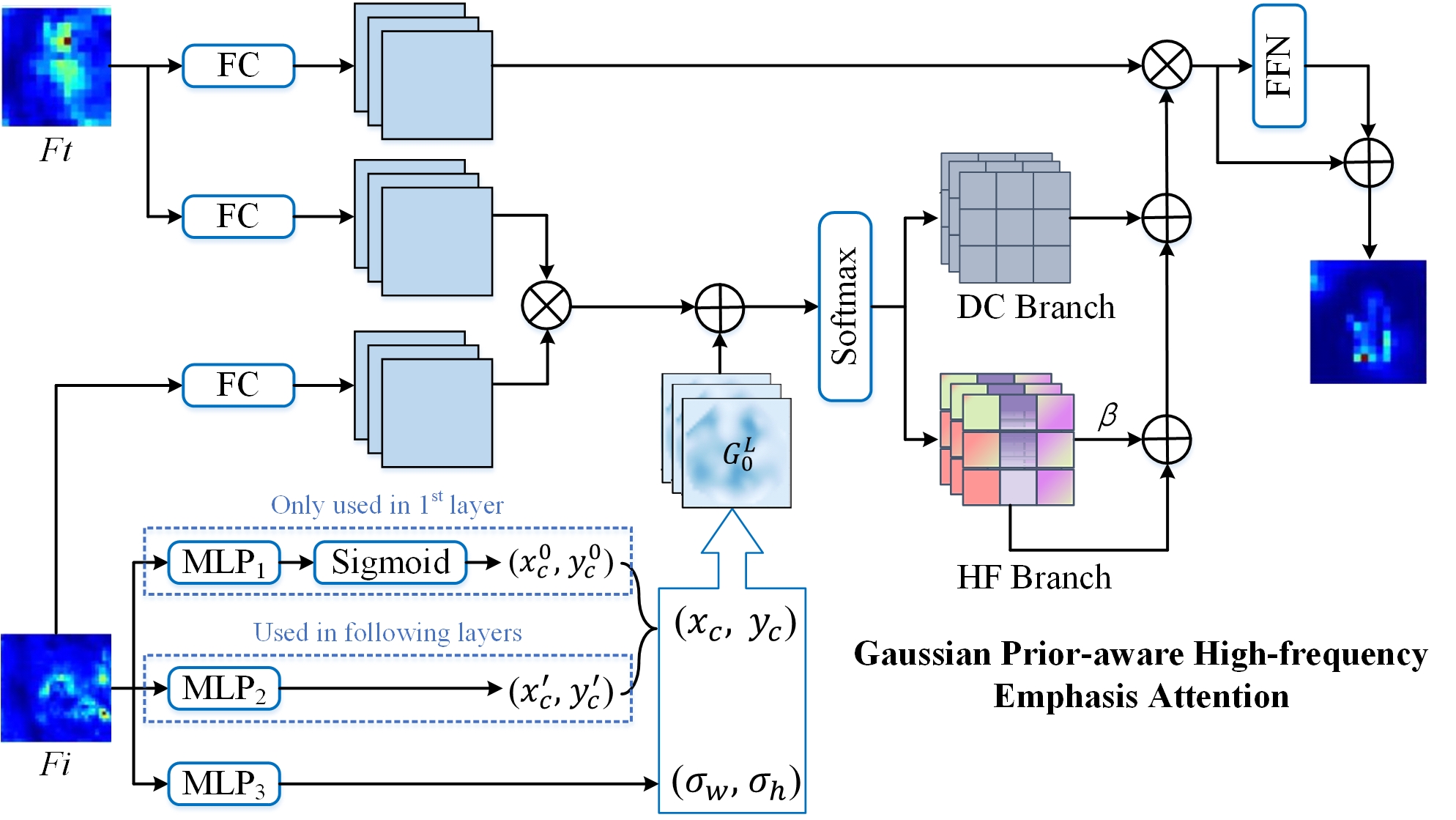} 
\caption{ One-layer detailed architecture of our proposed Spatial-Frequency Transformer.}  
\label{attenGPHA}
\end{figure}

\noindent
\textbf{Gaussian Spatial Prior Attention. }
As shown in Fig.~\ref{attenGPHA}, the first key component of our proposed unified Spatial-Frequency Transformer is the Gaussian spatial prior generation. As we all know, the center position $(x_c, y_c)$ and standard deviation $(\sigma_w, \sigma_h)$ are two important parameters for the Gaussian-prior distribution generation. In this work, we adopt a simple but effective Gaussian Generation Network (GGN) to generate these parameters and variables adaptively. Specifically, the GGN contains three parallel MLP branches $MLP_1, MLP_2, MLP_3$ separately. The $MLP_1$ is the center prediction network, which is solely used in the first attention layer of SFFormer. The $MLP_2$ is used to predict the center position bias, which is deployed on the subsequent layers of SFFormer. Note that the center bias will fine-tune the initial center position from $MLP_1$ to adopt the specific situation of the current layer. The $MLP_3$ branch generates the variance of weight and height of Gaussian weight, also called the standard deviation of $G(x, y)$.

Given the search features $F_i$, we can get the Gaussian map using GGN, and the detailed procedure can be written as: 
\begin{equation}
\begin{aligned}
(x^0_c, y^0_c) &= Sigmoid(MLP_1(F_i)) \\
(x'_c, y'_c) &= MLP_2(F_i), \ (\sigma_w, \sigma_h) = MLP_3(F_i) \\
(x_c, y_c) & = (x^0_c, y^0_c) + (x'_c, y'_c) \\
G(x, y) & = \exp \left(-\alpha\left(\frac{\left(x-x_{c}\right)^{2}}{2 \sigma_{w}^{2}}+\frac{\left(y-y_{c}\right)^{2}}{2 \sigma_{h}^{2}}\right)\right)
\end{aligned}
\end{equation}
where $\alpha$ is the hyper-parameter to control the scaling factor. As a deformation of standard normal distribution, Gaussian distribution has the same characteristic, i.e., assign the patches with different weight factors according to the distance with the center position. Therefore, Gaussian distribution has innate preference weight with patch positions.

After we obtained the Gaussian prior maps, the second question is how to integrate them into the attention mechanism. Inspired by the fact that Gaussian maps are also 2D prior images, we treat them as bias terms and fuse them into the similarity matrix of the self-attention mechanism. This procedure can be formulated as: 
\begin{equation}
\begin{aligned}
W_{Attn}=\operatorname{Softmax}\left(\frac{\mathcal{Q} \mathcal{K}^{\mathrm{T}}}{\sqrt{d_{k}}} + log G(x, y) \right)
\end{aligned}
\end{equation}
where $log$ denotes the logarithm operation for smooth attention decay of Gaussian map, $W_{Attn}$ represents the weight of query attention, and $d_{k}$ is the dimension of key feature vectors.

To better fit the Gaussian prior maps with the multi-head attention mechanism, we extend the aforementioned Gaussian maps into multi-head fields. Specifically, we broaden the output of single-head Gaussian map generation network $MLP_1, MLP_2, MLP_3$ dimensions from $2, 2, 2$ into $2\times N, 2\times N, 2\times N$, where $N$ is the head number of multi-head attention. Then, the proposed multi-head Gaussian prior attention weight can be rewritten as: 
\begin{equation}
\begin{aligned}
W^n_{Attn}=\operatorname{Softmax}\left(\frac{\mathcal{Q}_n \mathcal{K}_n^{\mathrm{T}}}{\sqrt{d_{k}}} + log G_n(x, y) \right)
\end{aligned} \label{eq6}
\end{equation}
where $n \in [1,\dots, N]$.  
Based on the Gaussian-weighted attention weight, our Gaussian Prior Attention (GPA) module can be formulated as: 
\begin{equation}
\begin{aligned}
H_n & = W^n_{Attn}\cdot \mathcal{V}_n \\
GPA(\mathcal{Q}, \mathcal{K}, \mathcal{V}, G) & = FFN[H_{1}, \ldots, H_{N}] 
\end{aligned} 
\end{equation}
where the $FFN$ consists of three FC layers.

\noindent
\textbf{High-Frequency Emphasis Attention. }
Another key component of our SFFormer is the High-Frequency Emphasis Attention which significantly enhances the low-/high-frequency feature learning. It is inspired by the recent findings that the attention mechanism inherently amounts to a low-pass filter~\cite{nt2019revisiting, cai2020note}. Therefore, the deep stacked multi-head attention layers in the Transformer severely suppress the high-frequency component of input features and generate over-smoothing output feature sequences. This may lead to the degeneration towards Direct-Current (DC) component~\footnote{\url{https://en.wikipedia.org/wiki/Direct_current}}. To protect the high-frequency features meanwhile keep the DC component, we decompose the co-attention weight into DC and high-frequency components. 

As shown in Fig.~\ref{attenGPHA}, given the Gaussian-weighted similarity matrix $W^n_{Attn} \in \mathbb{R}^{N\times S_I\times S_T}$, we first decompose it into dual branches: the average DC bias matrix branch and high-frequency components branch. 
More in detail, this procedure can be formulated as: 
\begin{equation}
\begin{aligned}
W^n_{Attn\_dc} &= \frac{\textbf{1}}{Len(\mathcal{V})}  \\
W^n_{Attn\_hf} &= W^n_{Attn} - W^n_{Attn\_dc} \\
W^n_{Attn\_hf} &= W^n_{Attn\_hf} \times (1 + \beta_n) \\
\hat{W}^n_{Attn} &= W^n_{Attn\_dc} + W^n_{Attn\_hf}
\end{aligned}
\end{equation}
where $ \textbf{1} \in \mathbb{R}^{N\times S_I\times S_T}$. 
As illustrated in Eq.~\ref{eq6}, for each query patch, the sum of similarity scores, i.e., the attention weights, between a query patch and value sequence is always one due to the use of $Softmax$ operator. Therefore, the DC component of the similarity matrix $W^n_{Attn}$ is averaged by the value sequence length. 
Correspondingly, the high-frequency signal can be obtained by subtracting DC bias from the given attention weight matrix. To emphasize the high-frequency attention weight, we introduce a learnable parameter vector $\beta \in \mathbb{R}^{1\times N}$ into the high-frequency branch. $\beta_n$ is initialized with zero and assigned as the specific and adaptive magnification of the $n^{th}$ separated attention head.

Through the high-frequency emphasis attention module, we evolve the attention from the low-pass filter into an all-pass filter to avoid the degradation of the high-frequency feature. 
Note that, emphasizing the attention weight component will not affect the basic attention flow structure and merely introduce $N\times L$ extra parameters. $N$ and $L$ denote the number of attention heads and layers used in SFFormer. Therefore, the final formulation of our newly proposed Gaussian Prior aware High-frequency emphasis Attention (GPHA) can be written as: 
\begin{equation}
\begin{aligned} 
\label{eq9} 
\hat{H}_n &= \hat{W}^n_{Attn} \cdot \mathcal{V} \\
GPHA(\mathcal{Q}, \mathcal{K}, \mathcal{V}, G) &= FFN[\hat{H}_{1}, \ldots, \hat{H}_{N}] 
\end{aligned} 
\end{equation}  
In our practical implementation, this module is stacked for $L$ layers for building SFFormer and our visual tracker. The outcome of the last GPHA block will be directly fed into classification and regression heads for target localization. To simplify and keep a clean framework, we adopt dual simple and classical three FC layers as the tracking heads.

\subsection{Training and Tracking Phase}  \label{train_test_phase}

\noindent
\textbf{Training Phase. }  
Given two frames in a random sampling within a 200-frame window video as template and search frame, the training framework will resize the template and search frame into $128\times 128, 256\times 256$ respectively. The proposed tracker is optimized in a two-branch end-to-end fashion with three loss functions. For the classification prediction, we adopt the binary cross-entropy (BCE) loss which can be formulated as:
\begin{equation}
\begin{aligned}
L_{b c e}(y,y^{\prime})=\sum_{s=1}^{S_I} -y_{s} \log y_{s}^{\prime}
\end{aligned}
\end{equation}
where $s$ is the $s^{th}$ patch in the predicted probability sequence. $ y_{s}$, $y_{s}^{\prime}$ separately denote the ground truth label and predicted class.

Different from existing works which adopt the rectangle classification, in this work, we employ the ellipse ground-truth label which can be written as: 
\begin{equation}
\begin{aligned}
y_n =\left\{\begin{array}{ll}
0, & if~~\frac{\left(p_{x}-g_{x_{c}}\right)^{2}}{\left(\frac{g_{w}}{2}\right)^{2}}+\frac{\left(p_{y}-g_{y_{c}}\right)^{2}}{\left(\frac{g_{h}}{2}\right)^{2}} > 1 \\
1, & otherwise 
\end{array}\right.
\end{aligned}
\end{equation}
where $(g_{x_{c}}, g_{y_{c}}, g_{w}, g_{h})$ denote the center coordinate, width, and height of ground-truth boxes, $(p_{x},p_{y})$ is the classified point. When the point falls in the ellipse label, we regard it as the foreground class. Otherwise, outside the ellipse is the background. 
Inspired by the Hungarian algorithm in DETR\cite{detr}, we combine the CIOU loss~\cite{ciou} and L1 loss for the position prediction. The regression loss can be written as: 
\begin{equation}
\begin{aligned}
L_{reg}(b_{s}, \hat{b}_{s})=\lambda_{1} L_{\text {ciou }}\left(b_{s}, \hat{b}_{s}\right)+\lambda_{2} L_{1}\left(b_{s}, \hat{b}_{s}\right)
\end{aligned}
\end{equation}
where $b_{s}, \hat{b}_{s}$ shows the ground-truth box and prediction box coordinate. $\lambda_{1}$ and $\lambda_{2}$ are the weight hyper-parameters to balance the CIOU and L1 loss. 
Therefore, the overall loss function can be written as: 
\begin{equation}
\begin{aligned}
L = \lambda_{1} L_{\text {ciou }}\left(b_{s}, \hat{b}_{s}\right)+\lambda_{2} L_{1}\left(b_{s}, \hat{b}_{s}\right) + \lambda_{3} L_{b c e}(y,y^{\prime})  
\end{aligned}
\end{equation}
We simply set the trade-off parameter $\lambda_{1}=5$, $\lambda_{2}=2$, $\lambda_{3}=10$ without any hyper-parametric searching.

\noindent
\textbf{Tracking Phase. }
During inference, we crop the template size from the first frame in the video sequence and feed it into our backbone network to get the template feature. The template feature will be cached and employed in the following tracking phase. For the second and subsequent frames, we crop them into $256\times 256$ search region. Then, the template and search feature sequences are fed into the end-to-end framework in parallel to gain classification and regression maps.

To achieve robust long-term visual tracking, in this work, we adopt a simple but effective local-global search strategy. It is inspired by~\cite{tanet} but much simplified to avoid extra parameters and computational cost. The major difference between local and global search is shown below: 
\begin{itemize}
\item Local search strategy: employing the center and four times area of the previous frame bounding box as the search region of search frames. 
\item Global search strategy: using the whole image as the search region of search frames. 
\end{itemize}

Because our tracker crops the search region into $256\times 256$ immediately after selecting the search region, the global strategy will keep the same computational cost as the local and have not any hyper-parameters introduced. In our tracker, the switching from local to global search will be executed when the classification confidence is less than the threshold $\tau_c$ for $\tau_f$ consecutive frames. Inspired by the local-global switch strategy proposed in TANet~\cite{tanet}, we simply set $\tau_f=5$. To achieve a good balance between robustness and high-performance tracking, we simply set the failure threshold as $\tau_c=0.98$. Note that these two parameters can be optimized by a hyper-parametric searching scheme. We leave this in future work.

\section{Experiments}
\subsection{Datasets and Evaluation Metrics} 
For the training of our model, we randomly sample training pairs from COCO~\cite{coco}, TrackingNet~\cite{trackingnet}, LaSOT~\cite{lasot}, and GOT-10K~\cite{got10k} datasets. 
To evaluate the effectiveness of our proposed tracker SFTransT, we test and verify our model on nine tracking datasets, including OTB100~\cite{otb100}, NFS~\cite{nfs}, GOT-10K~\cite{got10k}, UAV123~\cite{uav123}, TNL2K~\cite{TNL2K}, TrackingNet~\cite{trackingnet}, WebUAV-3M~\cite{webuav}, LaSOT~\cite{lasot}, and LaSOT\_ext\cite{lasotext}. Note that the last two are long-term tracking datasets.

In our experiments, multiple evaluation metrics are adopted, including \textbf{Success Rate}, \textbf{Precision Rate}, \textbf{Normalized Precision Rate}, and \textbf{Average Overlaps}. Specifically, 
Success Rate (SR) is the metric used to measure the IOU overlap between ground-truth and prediction boxes.  
Precision Rate (PR) measures the center's precise degree between the predicted point and ground-truth center. Usually, it will be treated as a successful track when the distance is lower than 20 pixels.
Normalized Precision Rate (NPR) extra considers the scale size of the ground truth box and normalizes the precision rate. 
SR and PR are the all-purpose measure metrics in OTB100, NFS, UAV123, LaSOT, TrackingNet, LaSOT\_ext, WebUAV-3M and TNL2K, while NPR is usually used in TrackingNet, LaSOT, LaSOT\_ext, WebUAV-3M, and TNL2K. 
Additionally, the Average Overlaps (AO), SR$_{0.5}$, and SR$_{0.75}$ are specific metrics for GOT-10K. AO calculates the average overlaps between all ground-truth and predicted bounding boxes. SR$_{0.5}$ and SR$_{0.75}$ laid down the different thresholds in the IOU overlap of SR.

\begin{table*}[!htp]
\centering
\caption{Performance comparison on the GOT-10K benchmark. The best results are shown in \textbf{bold}. }
\resizebox{\textwidth}{!}{
\begin{tabular}{l|llllllllll}
\toprule
& \multicolumn{1}{c}{SiamRPN++}  & \multicolumn{1}{c}{Ocean} & \multicolumn{1}{c}{SiamR-CNN}  & \multicolumn{1}{c}{TransT}  & \multicolumn{1}{c}{STARK}  & \multicolumn{1}{c}{DTT} & \multicolumn{1}{c}{SwinTrack}  & \multicolumn{1}{c}{SBT}   & \multicolumn{1}{c}{MixFormer} & \multicolumn{1}{c}{\textbf{Ours}}  \\
& \multicolumn{1}{c}{~\cite{dimp}}  & \multicolumn{1}{c}{~\cite{ocean}} & \multicolumn{1}{c}{~\cite{siamr-cnn}} & \multicolumn{1}{c}{~\cite{transt}} &\multicolumn{1}{c}{-ST101\cite{stark}} &  \multicolumn{1}{c}{~\cite{dtt}}  &  \multicolumn{1}{c}{-T~\cite{swintrack}} &  \multicolumn{1}{c}{-large\cite{sbt}}   & \multicolumn{1}{c}{-1k\cite{mixformer}}  \\   
\midrule
AO ($\%$)    & \multicolumn{1}{c}{51.7} & \multicolumn{1}{c}{61.1}   & \multicolumn{1}{c}{64.9}   & \multicolumn{1}{c}{67.1}  & \multicolumn{1}{c}{68.8}  & \multicolumn{1}{c}{68.9} &\multicolumn{1}{c}{69.0} &\multicolumn{1}{c}{70.4} &\multicolumn{1}{c}{71.2} & \multicolumn{1}{c}{\textbf{72.7}}\\
SR$_{0.5}$ ($\%$) & \multicolumn{1}{c}{61.6} & \multicolumn{1}{c}{72.1} & \multicolumn{1}{c}{72.8}   & \multicolumn{1}{c}{76.8} &\multicolumn{1}{c}{78.1}  & \multicolumn{1}{c}{79.8} &\multicolumn{1}{c}{78.1}  &\multicolumn{1}{c}{80.8}  &\multicolumn{1}{c}{79.9}    & \multicolumn{1}{c}{\textbf{84.3}}  \\
SR$_{0.75}$ ($\%$)    & \multicolumn{1}{c}{32.5} & \multicolumn{1}{c}{47.3}  & \multicolumn{1}{c}{59.7}   & \multicolumn{1}{c}{60.9} &\multicolumn{1}{c}{64.1}  & \multicolumn{1}{c}{62.2}  &\multicolumn{1}{c}{62.1}    &\multicolumn{1}{c}{64.7}     &\multicolumn{1}{c}{65.8}     & \multicolumn{1}{c}{\textbf{66.9}} \\ 
\textcolor{black}{Speed (FPS)} &\multicolumn{1}{c}{\textcolor{black}{35}}    &\multicolumn{1}{c}{ \textcolor{black}{25} }    &\multicolumn{1}{c}{ \textcolor{black}{5}}    &\multicolumn{1}{c}{ \textcolor{black}{50}}    &\multicolumn{1}{c}{\textcolor{black}{32}}    &\multicolumn{1}{c}{ \textcolor{black}{50} }    &\multicolumn{1}{c}{ \textcolor{black}{\textbf{98}} }    &\multicolumn{1}{c}{ \textcolor{black}{24} }    &\multicolumn{1}{c}{ \textcolor{black}{25}}    &\multicolumn{1}{c}{\textcolor{black}{27}}     \\
\textcolor{black}{Params (M)} &\multicolumn{1}{c}{\textcolor{black}{54}}    &\multicolumn{1}{c}{ \textcolor{black}{-} }    &\multicolumn{1}{c}{ \textcolor{black}{-}}    &\multicolumn{1}{c}{ \textcolor{black}{\textbf{23}}}    &\multicolumn{1}{c}{\textcolor{black}{42.9}}    &\multicolumn{1}{c}{ \textcolor{black}{-} }    &\multicolumn{1}{c}{ \textcolor{black}{23} }    &\multicolumn{1}{c}{ \textcolor{black}{-} }    &\multicolumn{1}{c}{ \textcolor{black}{35.6}}    &\multicolumn{1}{c}{\textcolor{black}{29.6}}     \\
\bottomrule
\end{tabular}
}
\label{got10ktable}
\end{table*}

\begin{table*}[!htp]
\centering
\caption{Performance comparison on the TrackingNet benchmark. The best results are shown in \textbf{bold}. }
\resizebox{\textwidth}{!}{
\begin{tabular}{l|llllllllll}
\toprule
& \multicolumn{1}{c}{SiamFC++}  & \multicolumn{1}{c}{PrDiMP} & \multicolumn{1}{c}{STMTracker}  & \multicolumn{1}{c}{SiamR-CNN}  & \multicolumn{1}{c}{STARK}   & \multicolumn{1}{c}{SwinTrack} & \multicolumn{1}{c}{TransT} & \multicolumn{1}{c}{ToMP101}   & \multicolumn{1}{c}{MixFormer} & \multicolumn{1}{c}{\textbf{Ours}}  \\  
& \multicolumn{1}{c}{~\cite{siamfc++}}  & \multicolumn{1}{c}{~\cite{prdimp}} & \multicolumn{1}{c}{~\cite{stmtrack}} & \multicolumn{1}{c}{~\cite{siamr-cnn}} &\multicolumn{1}{c}{-ST101\cite{stark}} &  \multicolumn{1}{c}{-T\cite{swintrack}}  &  \multicolumn{1}{c}{\cite{transt}} &  \multicolumn{1}{c}{\cite{tomp}}   & \multicolumn{1}{c}{-1k\cite{mixformer}}  \\ 
\midrule
SR ($\%$)    & \multicolumn{1}{c}{75.4}  & \multicolumn{1}{c}{75.8}  & \multicolumn{1}{c}{80.3}  &  \multicolumn{1}{c}{81.2}   &  \multicolumn{1}{c}{82.0}    &  \multicolumn{1}{c}{80.8}  &\multicolumn{1}{c}{81.4}   &  \multicolumn{1}{c}{81.5}        &  \multicolumn{1}{c}{82.6}         & \multicolumn{1}{c}{\textbf{82.9}} \\
NPR ($\%$) & \multicolumn{1}{c}{80.0}   & \multicolumn{1}{c}{81.6}      & \multicolumn{1}{c}{85.1}   & \multicolumn{1}{c}{85.4}      & \multicolumn{1}{c}{86.9}       & \multicolumn{1}{c}{85.5}    & \multicolumn{1}{c}{86.7}  & \multicolumn{1}{c}{86.4}       & \multicolumn{1}{c}{\textbf{87.7}}   & \multicolumn{1}{c}{87.3}    \\
PR ($\%$)   & \multicolumn{1}{c}{70.5}   & \multicolumn{1}{c}{70.4}      & \multicolumn{1}{c}{76.7}    & \multicolumn{1}{c}{80.0}    & \multicolumn{1}{c}{-}   & \multicolumn{1}{c}{77.9}   & \multicolumn{1}{c}{80.3}   & \multicolumn{1}{c}{78.9}   & \multicolumn{1}{c}{81.2}     & \multicolumn{1}{c}{\textbf{81.3}}    \\ \bottomrule
\end{tabular}
}
\label{trackingnettable}
\end{table*}

\subsection{Implementation Details} 
The used backbone Swin-Tiny is initialized using pre-trained weights on the ImageNet-1K dataset. For all of the attention modules in our framework, we set the dropout as 0.1. Sampled image pairs are cropped with brightness jittering and horizontal flips. The training epoch, iteration, and mini-batch are 500, 1k, and 48, respectively. We set the initial learning rate as 1e-5 for optimizing the cross-scale fusion-based Swin-Tiny and 1e-4 for all others without pre-trained components. The optimizer is AdamW~\cite{adamw} with a weight decay of 1e-4. The learning rate has decayed a magnitude at the 400$^{th}$ epoch. The SFFormer consist of six GPHA blocks. Our code is implemented based on Python 3.8 and Pytorch 1.8.0 on a server with $4\times$ RTX 3090 GPUs. 
\textcolor{black}{
The total training time of 500 epochs on four datasets is around 70 hours, while the inference time of each frame is 0.037 ms (27 frames per second) on a single RTX3090. 
}

\subsection{Comparison with Other State-of-the-art Trackers}    

In this section, we report our tracking performance on nine widely used benchmark datasets and compare it with other state-of-the-art (SOTA) trackers.

\noindent 
\textbf{Results on OTB100~\cite{otb100} and NFS~\cite{nfs}}: OTB100 is a short-term and approaching saturation benchmark used in visual tracking. The performance comparison can be found in Table~\ref{otbnfs}, most trackers achieve SR results of around 70\%. Our SFTransT obtains a slightly better result compared to ToMP101, with 70.3\%/91.6\% on SR/PR. 
For the NFS~\cite{nfs} dataset, the 30 FPS (Frame Per Second) version is adopted for evaluation which is full of fast motions and challenging distractors. As shown in Table~\ref{otbnfs}, our tracker achieves a performance of 66.0\%/80.3\% on SR/PR, which is slightly better than TransT~\cite{transt} and comparable with ToMP101.

\begin{table}[!htp]
\centering
\caption{Success plot comparison on OTB and NFS benchmark. The best results are shown in \textbf{bold}.}
\resizebox{\columnwidth}{!}{
\begin{tabular}{l|lllllllll}  \toprule
& \multicolumn{1}{c}{SiamR-CNN} & \multicolumn{1}{c}{PrDiMP} & \multicolumn{1}{c}{STARK-ST50}  & \multicolumn{1}{c}{TransT} & \multicolumn{1}{c}{MixFormer-1k}  & \multicolumn{1}{c}{ToMP101}     & \multicolumn{1}{c}{\textbf{Ours}}  \\  
& \multicolumn{1}{c}{~\cite{siamr-cnn}}  & \multicolumn{1}{c}{~\cite{prdimp}} & \multicolumn{1}{c}{~\cite{stark}} & \multicolumn{1}{c}{~\cite{transt}} &\multicolumn{1}{c}{\cite{mixformer}} &  \multicolumn{1}{c}{\cite{tomp}}  &  \multicolumn{1}{c}{}  \\
\midrule
OTB100 &\multicolumn{1}{c}{70.1}   &\multicolumn{1}{c}{69.6}     &\multicolumn{1}{c}{68.5}   &\multicolumn{1}{c}{69.4}   &\multicolumn{1}{c}{69.6} &\multicolumn{1}{c}{70.1}  &\multicolumn{1}{c}{\textbf{70.3}}   \\
NFS    &\multicolumn{1}{c}{63.9}  &\multicolumn{1}{c}{63.5}    &\multicolumn{1}{c}{65.2}    &\multicolumn{1}{c}{65.7}     &\multicolumn{1}{c}{-}  &\multicolumn{1}{c}{\textbf{66.9}}  &\multicolumn{1}{c}{66.0}    \\ \bottomrule  
\end{tabular}
} \label{otbnfs}
\end{table}

\noindent
\textbf{Results on GOT-10K~\cite{got10k}}: GOT-10K is a recently released large-scale short-term benchmark in visual tracking.  Following the one-shot training protocols of GOT-10K, our tracker merely use the training set of GOT-10K for model learning. We evaluate our testing results on the benchmark official online server. As shown in Table~\ref{got10ktable}, our tracker achieves the new SOTA performance with the 72.7\%/84.3\%/66.9\% on the AO/SR$_{0.50}$/SR$_{0.75}$, surpassing the previous SOTA tracker MixFormer-1k by 1.5\%/4.4\%/1.1\%. 
Compared with the one-stage one-branch tracker SBT-large (the largest version of SBT~\cite{sbt}), we overstep 2.3\% in AO. 
SwinTrack-T is a tracker with the same backbone network (Swin-Tiny~\cite{swin}) as ours. Our tracker outperforms it with a remarkable gap with our spatial-frequency design. 
Furthermore, compared with TransT~\cite{transt}, we also gained a significant 5.6\% performance improvement. 
\textcolor{black}{As for speed, we can find that our proposed SFTransT can run at 27 FPS, faster than the previous SOTA tracker MixFormer and SBT. It is also easy to find from Table~\ref{got10ktable}, our proposed SFTransT contains 29.6 Million (M) parameters, less than the previous SOTA method MixFormer-1k and comparable with SwinTrack and TransT. 
}

\noindent
\textbf{Results on TrackingNet~\cite{trackingnet}}: TrackingNet contains more than 30K training videos with a 30-frame interval labelling strategy. It covers an average of 1300 frames per sequence. As shown in Table~\ref{trackingnettable}, our tracker achieves an advanced performance on this large-scale benchmark dataset, i.e., 82.9\%, 87.3\%, and 81.3\% on the SR/NPR/PR metrics, respectively. More in detail, our results are better than recent strong Transformer based trackers, including ToMP101~\cite{tomp} (81.5\%, 86.4\%, 78.9\%), TransT~\cite{transt} (81.4\%, 86.7\%, 80.3\%).

\noindent
\textbf{Results on UAV123~\cite{uav123}}:  UAV123 is a dataset collected by the low-attribute intelligent unmanned aerial vehicle (UAV). It consists of 123 sequences with many long-term videos and small target objects inside. As shown in Table~\ref{uavtable}, our tracker achieves a new state-of-the-art result, especially 71.3\% on SR, outperforming MixFormer-1k by +2.6\%.

\begin{table}[!t]
\centering
\caption{Performance comparison on the UAV123 benchmark. The best results are shown in \textbf{bold}.}
\resizebox{\columnwidth}{!}{
\begin{tabular}{l|lllllll}
\toprule
& \multicolumn{1}{c}{SiamRPN++} & \multicolumn{1}{c}{SiamGAT} & \multicolumn{1}{c}{SiamR-CNN}  & \multicolumn{1}{c}{TransT}  & \multicolumn{1}{c}{MixFormer-1k} & \multicolumn{1}{c}{\textbf{Ours}}  \\
& \multicolumn{1}{c}{\cite{siamrpn++}}  & \multicolumn{1}{c}{\cite{siamgat}} & \multicolumn{1}{c}{\cite{siamr-cnn}} & \multicolumn{1}{c}{\cite{transt}} &\multicolumn{1}{c}{\cite{mixformer}} &  \multicolumn{1}{c}{} \\
\midrule
SR ($\%$)    &\multicolumn{1}{c}{61.0} &\multicolumn{1}{c}{64.6}   &\multicolumn{1}{c}{64.9}   &\multicolumn{1}{c}{69.1}  &\multicolumn{1}{c}{68.7}   &\multicolumn{1}{c}{\textbf{71.3}}\\
PR ($\%$)    &\multicolumn{1}{c}{80.3} &\multicolumn{1}{c}{84.3}   &\multicolumn{1}{c}{83.4}   &\multicolumn{1}{c}{-}    &\multicolumn{1}{c}{89.5}    &\multicolumn{1}{c}{\textbf{89.7}}     \\ \bottomrule
\end{tabular}}
\label{uavtable}
\end{table}

\begin{table*}[!htp]
\centering
\caption{Performance comparison on the LaSOT benchmark. The best results are shown in \textbf{bold}.}
\resizebox{\textwidth}{!}{
\begin{tabular}{l|lllllllllll}
\toprule
& \multicolumn{1}{c}{AutoMatch}  & \multicolumn{1}{c}{SiamR-CNN} & \multicolumn{1}{c}{TrDiMP} & \multicolumn{1}{c}{TransT}  & \multicolumn{1}{c}{CSWinTT}   & \multicolumn{1}{c}{SBT}  & \multicolumn{1}{c}{SwinTrack}  & \multicolumn{1}{c}{KeepTrack}   & \multicolumn{1}{c}{ToMP101} & \multicolumn{1}{c}{MixFormer} & \multicolumn{1}{c}{\textbf{Ours}}  \\
& \multicolumn{1}{c}{~\cite{automatch}}  & \multicolumn{1}{c}{~\cite{siamr-cnn}} & \multicolumn{1}{c}{~\cite{trdimp}} & \multicolumn{1}{c}{~\cite{transt}} &\multicolumn{1}{c}{\cite{cswintt}} &  \multicolumn{1}{c}{large~\cite{sbt}}  &  \multicolumn{1}{c}{-T\cite{swintrack}} &  \multicolumn{1}{c}{\cite{keeptrack}}   & \multicolumn{1}{c}{\cite{tomp}}  & \multicolumn{1}{c}{-1k~\cite{mixformer}} \\ 
\midrule
SR ($\%$) &\multicolumn{1}{c}{58.2} &\multicolumn{1}{c}{64.8}  &\multicolumn{1}{c}{63.7}   &\multicolumn{1}{c}{64.9}   &\multicolumn{1}{c}{66.2} &\multicolumn{1}{c}{66.7} &\multicolumn{1}{c}{66.7} &\multicolumn{1}{c}{67.1}   &\multicolumn{1}{c}{68.5} &\multicolumn{1}{c}{67.9} &\multicolumn{1}{c}{\textbf{69.0}}  \\
NPR ($\%$) &\multicolumn{1}{c}{-}    &\multicolumn{1}{c}{72.2}    &\multicolumn{1}{c}{-}      &\multicolumn{1}{c}{73.8}   &\multicolumn{1}{c}{75.2} &\multicolumn{1}{c}{-}  &\multicolumn{1}{c}{75.8}  &\multicolumn{1}{c}{77.2}   &\multicolumn{1}{c}{\textbf{79.2}} &\multicolumn{1}{c}{77.3}  &\multicolumn{1}{c}{78.1}   \\
PR ($\%$) &\multicolumn{1}{c}{59.9} &\multicolumn{1}{c}{68.4}   &\multicolumn{1}{c}{61.4}   &\multicolumn{1}{c}{69.0}   &\multicolumn{1}{c}{70.9} &\multicolumn{1}{c}{71.1} &\multicolumn{1}{c}{70.6}  &\multicolumn{1}{c}{70.2}   &\multicolumn{1}{c}{73.5} &\multicolumn{1}{c}{73.9} &\multicolumn{1}{c}{\textbf{73.9}}    \\  \bottomrule
\end{tabular}
} \label{lasottable}
\end{table*}

\begin{figure}
\centering
\includegraphics[width=1\columnwidth]{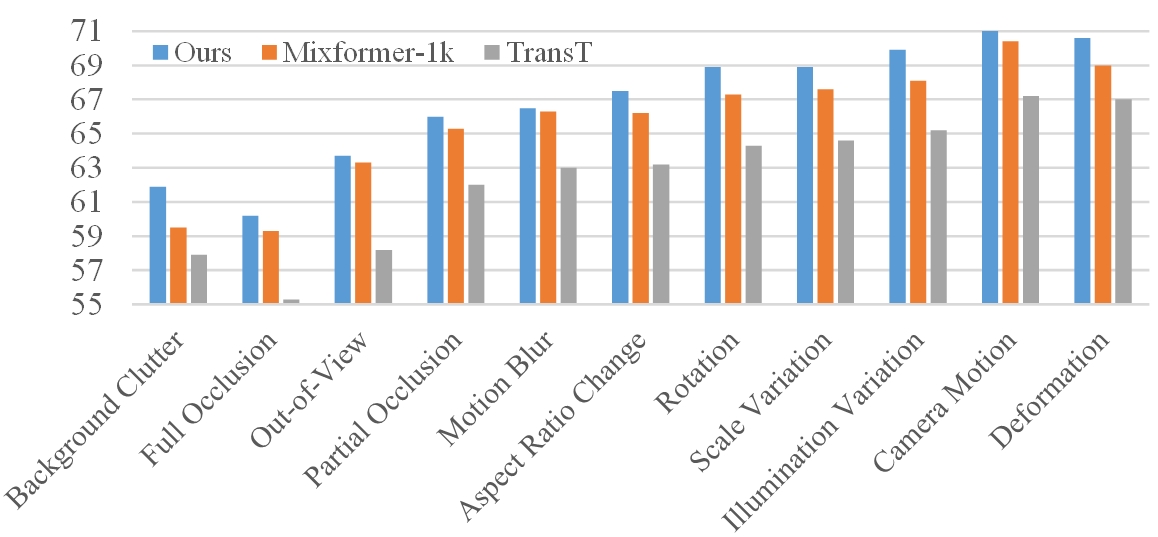} 
\caption{Attribute analysis on LaSOT~\cite{lasot} between Ours and
recent trackers Mixformer-1k and TransT.}
\label{lasot_attributes}
\end{figure}

\noindent 
\textbf{Results on LaSOT~\cite{lasot}}: As a long-term tracking benchmark, LaSOT contains an average of 2500 frames for the 280 testing videos. The experimental results are illustrated in Table~\ref{lasottable}, and many SOTA trackers are compared, e.g., MixFormer~\cite{mixformer}, ToMP~\cite{tomp}, SBT~\cite{sbt}, CSWinTT~\cite{cswintt}, KeepTrack~\cite{keeptrack}, SwinTrack~\cite{swintrack}. We can find that our proposed tracker achieves 69.0\%/78.1\% on SR/NPR, which is better than recent SOTA trackers SBT-large, ToMP101, SwinTrack-T, and Mixformer-1k. We also compare the SR scores on different attributes in Fig.~\ref{lasot_attributes}. Note that, our proposed tracker SFTransT outperforms the TransT by more than +4\% on all attributes.

\noindent 
\textbf{Results on LaSOT\_ext~\cite{lasotext}}: LaSOT\_ext is an extension of LaSOT benchmark dataset. It introduces 150 new video sequences with 15 new classes. It is also a long-term tracking dataset that contains over 2500 frames on average for each sequence. As shown in Table~\ref{lasotext}, our tracker achieves 46.4\% on SR and beats the previous SOTA tracker ToMP101.

\begin{table}[!htp]
\centering
\caption{Performance comparison on the LaSOT\_ext benchmark. The best results are shown in \textbf{bold}.}
\resizebox{\columnwidth}{!}{
\begin{tabular}{l|llllllll}
\toprule
& \multicolumn{1}{c}{SiamRPN++} & \multicolumn{1}{c}{DiMP} & \multicolumn{1}{c}{LTMU}  & \multicolumn{1}{c}{SuperDiMP} & \multicolumn{1}{c}{ToMP50} & \multicolumn{1}{c}{ToMP101}        & \multicolumn{1}{c}{\textbf{Ours}}  \\
& \multicolumn{1}{c}{~\cite{siamrpn++}}  & \multicolumn{1}{c}{~\cite{dimp}} & \multicolumn{1}{c}{~\cite{ltmu}} & \multicolumn{1}{c}{~\cite{dimp}} &\multicolumn{1}{c}{\cite{tomp}} &  \multicolumn{1}{c}{\cite{tomp}}  &  \multicolumn{1}{c}{}  \\
\midrule
SR ($\%$) &\multicolumn{1}{c}{34.0}  &\multicolumn{1}{c}{39.2}    &\multicolumn{1}{c}{41.4}  &\multicolumn{1}{c}{43.7}   & \multicolumn{1}{c}{45.4} & \multicolumn{1}{c}{45.9}  & \multicolumn{1}{c}{\textbf{46.4}}   \\
NP ($\%$) &\multicolumn{1}{c}{41.6}  &\multicolumn{1}{c}{47.6}   &\multicolumn{1}{c}{49.9}  &\multicolumn{1}{c}{56.3}     & \multicolumn{1}{c}{57.6}     & \multicolumn{1}{c}{\textbf{58.1}}    & \multicolumn{1}{c}{53.8} \\
PR ($\%$)  &\multicolumn{1}{c}{39.6}  &\multicolumn{1}{c}{45.1}   &\multicolumn{1}{c}{47.3}  &\multicolumn{1}{c}{-}   & \multicolumn{1}{c}{-}  & \multicolumn{1}{c}{-}    & \multicolumn{1}{c}{\textbf{54.1}}  \\  \bottomrule
\end{tabular}
} \label{lasotext}
\end{table}

\begin{figure*}
\centering
\includegraphics[width=1\textwidth]{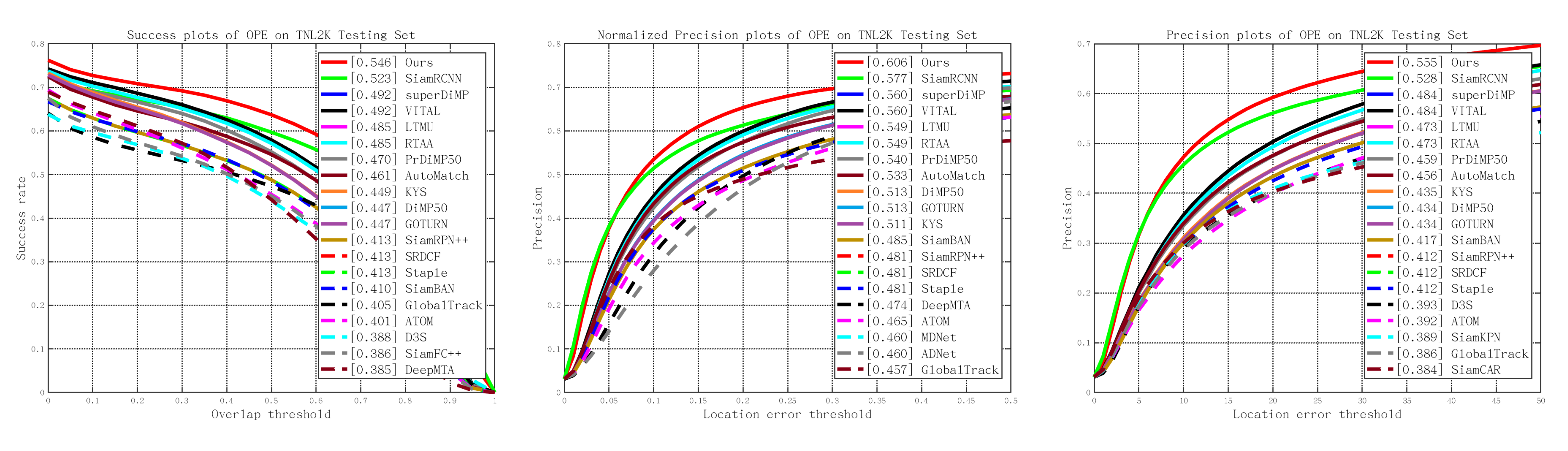} 
\caption{Comparison of our proposed tracker and other tracking models on the TNL2K dataset. Best viewed in color and zoom in. } 
\label{TNL2K_SR_PR_NPR}
\end{figure*}

\noindent 
\textbf{Results on TNL2K~\cite{TNL2K}}: TNL2K is a new high-quality benchmark  consisting of 700 high-diversity testing videos with natural language specifications. We follow the rule with bounding-box guided tracking and achieve the best performance on TNL2K. As shown in Fig.~\ref{TNL2K_SR_PR_NPR}, we receive state-of-the-art results with 54.6\%/60.6\%/55.5\% on SR/NPR/PR, respectively. This experiment demonstrates that our tracker obtains significant improvements, even compared with SOTA trackers such as SiamR-CNN~\cite{siamr-cnn}, SuperDiMP~\cite{dimp}, LTMU~\cite{ltmu}, AutoMatch~\cite{automatch}, KYS~\cite{kys}, etc.

\begin{figure*}
\centering
\includegraphics[width=1\textwidth]{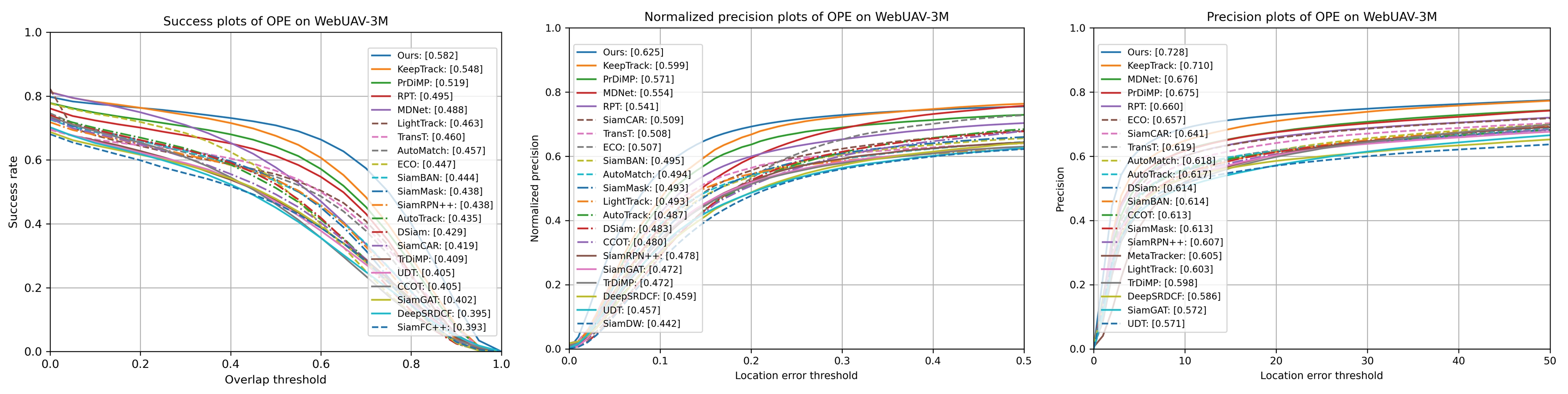} 
\caption{Comparison of our proposed tracker and other tracking models on the WebUAV-3M dataset. Best viewed in color and zoom in.} 
\label{webuav3m}
\end{figure*}

\noindent 
\textbf{Results on WebUAV-3M~\cite{webuav}}: WebUAV-3M is the newest and largest-scale UAV tracking benchmark to date, containing over 3.3 million frames across 4,500 videos with million-scale dense annotations. The test subset of WebUAV-3M has 780 videos and 120 target classes. We evaluate our SFTransT with the guidance of visual box annotations. As shown in Fig.~\ref{webuav3m}, SFTransT gets a new state-of-the-art result and surpasses KeepTrack (+3.4\%), TransT(+12.2\%) and AutoMatch(+12.5\%) on SR.

All in all, our proposed SFTransT achieves better performance on all these benchmarks than most of the compared trackers. We also set new state-of-the-art results on UAV123 and GOT-10K datasets, which fully validated the effectiveness and advantages of our tracker.

\subsection{Ablation Study}

In this section, extensive ablation studies are conducted to validate the effectiveness of our core components. 

\noindent
\textbf{Analysis on Feature Extraction, Fusion, and Augmentation.}  
In this subsection, we implement four kinds of feature extraction and fusion networks to validate the effectiveness of our feature extraction network, i.e., \#1, \#2, \#3, \#4, as shown in Table~\ref{ablation1}. More in detail, the Layer-wise Aggregation (LWA) module was first proposed in SiamRPN++, which linearly combines three output layers with learnable weights~\cite{siamrpn++, siamban}. \#4 represents our proposed Cross-Scale Fusion (CSF) based Swin Transformer feature extract network. 
ResNet50 is kept the same processing with SiamRPN++~\cite{siamrpn++} and SiamBAN~\cite{siamban}. 
As we can see from Table~\ref{ablation1}, when we replaced the backbone with ResNet50, the performances dropped around 1.1\%-1.8\% compared with the Swin-Tiny version. 
By comparing algorithm \#2 \textit{vs.} \#1, and algorithm \#4 \textit{vs.} \#3, we can find that the tracking results can be improved by more than 0.6\% in SR and 0.8\% in AO, with the help of CSF module. These experiments fully validated the effectiveness of our proposed cross-scale fusion based Swin-Tiny network.

In addition, we can also find that our feature interactive learning module MHCA helps for our final tracking results, by comparing algorithms \#4 and \#5. Specifically, this module improves the SR|PR, AO from $68.1|73.3$ to $69.0|73.9$, and $71.3$ to $72.7$ on the LaSOT and GOT-10K datasets, respectively.

\begin{table}[!htp] 
\caption{Ablation study about the feature extraction network and multi-scale fusion modules on long-term LaSOT and short-term GOT-10K benchmark.}
\centering
\begin{tabular}{l|l|c|c|c|cc} 
\toprule
\#  & Backbone   & Fusion &MHCA   & LaSOT  & GOT-10K  \\  
\midrule
1 & ResNet50   & LWA   &    & 65.7|69.9      &   69.0   \\
2 & ResNet50   & CSF  &    &  66.3|70.9      &   69.8    \\
3 & Swin-Tiny  & LWA  &    &  66.8|70.7      &    70.1     \\
4 & Swin-Tiny  & CSF  &    &  68.1|73.3       &    71.3    \\
5 & Swin-Tiny  & CSF  &$\checkmark$   &  69.0|73.9        & 72.7   \\
 \bottomrule     
\end{tabular}
\label{ablation1}
\end{table}

\begin{figure*}[!htp]
\centering
\includegraphics[width=1\textwidth]{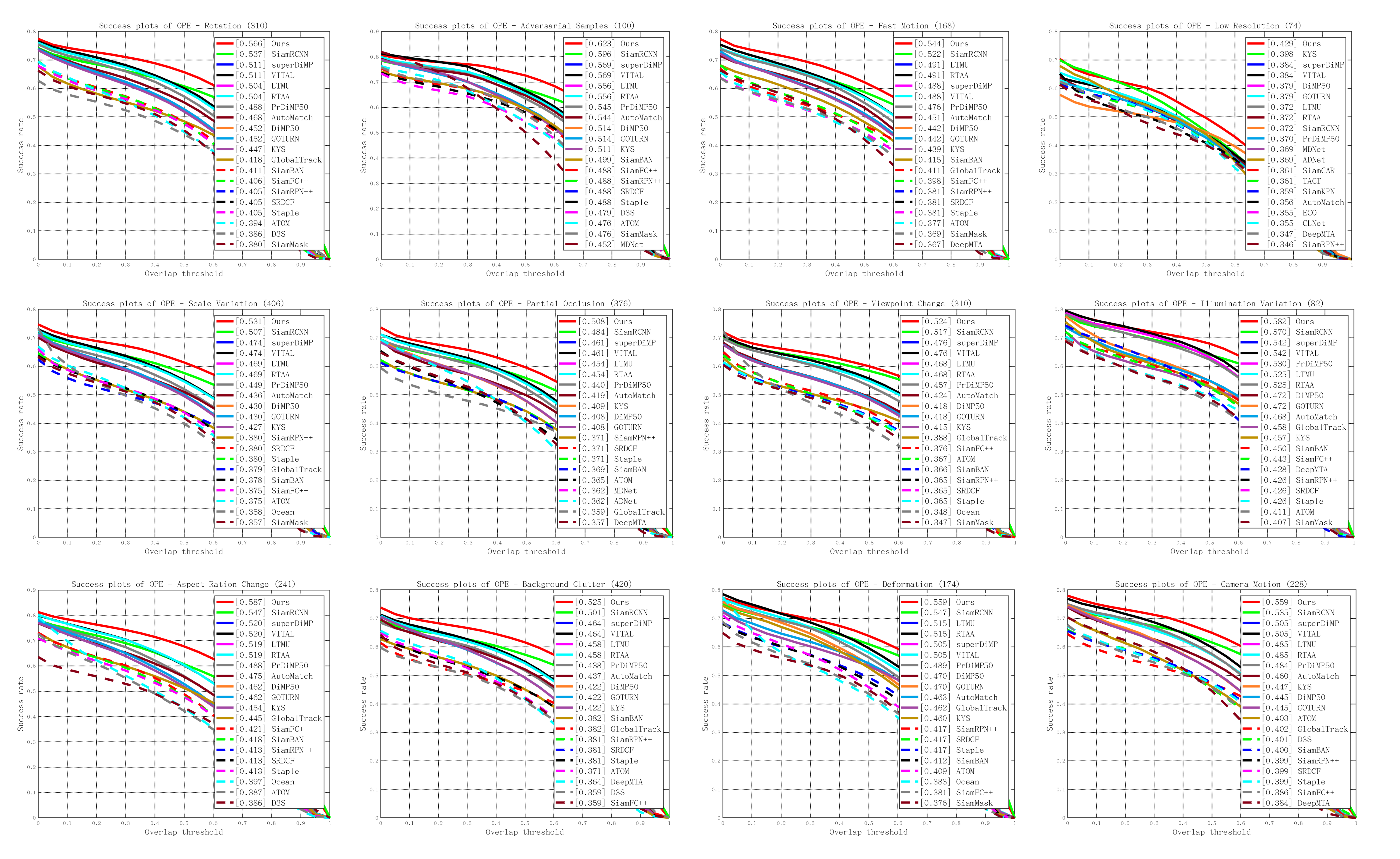} 
\caption{Tracking results under each challenging attribute on TNL2K dataset. Best viewed in color and zoom in.} 
\label{TNL2K_attributes}
\end{figure*}

\begin{table}[!htp] 
\caption{Ablation study of the GPHA component on the LaSOT and GOT-10K dataset. The (SR, PR) and (AO, SR$_{0.5}$, SR$_{0.75}$) are reported for the two datasets, respectively.}  
\resizebox{\columnwidth}{!}
{
\begin{tabular}{l|ll|llll}
\toprule
\# & Extraction          & LaSOT    & GOT-10K  \\ 
\hline 
1 & \ Baseline           & 65.9 | 70.3        & 67.7 | 77.9 | 62.4   \\
2 & + GPA w/o multi-G    & 66.5 | 70.9        & 68.1 | 79.3 | 63.0   \\
3 & + GPA w/ multi-G     & 68.0 | 72.1        & 70.9 | 81.8 | 65.7    \\
4 & + HEA w/o GPA        & 66.2 | 70.8        & 68.5 | 79.4 | 63.6    \\
5 & + GPHA               &\textbf{69.0} | \textbf{73.9}     &\textbf{72.7} | \textbf{84.3} | \textbf{66.9} \\
\bottomrule
\end{tabular}
}
\label{ablation2}
\end{table}

\noindent
\textbf{Analysis on Spatial-Frequency Transformer.}  \label{Analysis on Spatial-Frequency Transformer} 
To verify the effectiveness of our proposed SFFormer, multiple components are tested in this section, as shown in Table~\ref{ablation2}. Specifically, the \textbf{GPA} means the Gaussian-prior attention, \textbf{HEA} denotes the high-frequency emphasis attention module, \textbf{GPHA} is our proposed Gaussian-prior aware high-frequency emphasis attention. \textbf{w/o multi-G} means only using the same Gaussian maps for different attention heads in a GPA block. 

Algorithm \#1 listed in Table~\ref{ablation2} is the baseline tracker, which employs six-layer original attention only, without any extra spatial and frequency components. 
Comparing algorithms \#1 and \#2, it is easy to find that even simple Gaussian-prior attention can boost the tracking results on both datasets. Specifically, the baseline results are improved by +0.6|0.6, +0.4|1.4|0.6 on LaSOT (SR | PR) and GOT-10K (AO | SR$_{0.5}$ | SR$_{0.75}$) datasets.
When multi-G is adopted (i.e., tracker \#3), the overall performance can be improved from $65.9 | 70.3$ to $68.0 | 72.1$ on LaSOT, $67.7 | 77.9 | 62.4$ to $70.9 | 81.8 | 65.7$ on GOT-10K dataset. These results fully validated the effectiveness and advantages of our proposed Gaussian spatial prior for visual object tracking.

\textcolor{black}{ By comparing algorithm \#4 with \#1, we can see that the overall performance can be improved on both datasets, demonstrating the effectiveness of High-frequency Emphasis Attention (HEA). Note that, the improvement is less than 1\% on LaSOT dataset. This may be caused by the fact that the LaSOT is a long-term tracking benchmark dataset which is usually by many spatial challenges, e.g., frequently out-of-view and large-scale motion range. Its overall performance is influenced by many factors, in addition to the feature representation we handled in this paper. For the tracking results on the short-term GOT-10K dataset, we can find that the results can also be improved (+0.8\%|+1.5\%|+1.2\%) by introducing the HEA. It demonstrates that High-frequency Emphasis Attention (HEA) is effective in our framework and can not significantly improve the final performance solely. 
When combining the HEA with GPA, i.e., our proposed GPHA (\#5), the performance improves significantly by +1.0\%|+1.8\% on LaSOT and +1.8\%|+2.5\%|+1.2\% on GOT-10K compared with the solely GPA (\#3). It demonstrates that frequency emphasis can play a more significant role in visual tracking when combing with spatial attention. Furthermore, comparing our GPHA with the original self-attention module (\#5 vs. \#1), our proposed GPHA can improve the tracking by over 3\% on LaSOT and 5\% on GOT-10K. }
All of these results fully demonstrate the necessity and importance of spatial prior and frequency information for Transformer based tracking.

\noindent
\textbf{Analysis on Loss Functions and Classification Labels. } 
In this paper, we adopt different loss functions and classification labels for the training of our tracker. As shown in Table~\ref{ablation4}, we report and compare the tracking results with different settings on the training, validation, and testing subset of GOT-10K dataset. It is easy to find that the CIOU loss improves our model from 71.5 to 72.0 on the AO metric by comparing tracker \#1~\textit{vs.}~\#2. From the comparison of tracker \#1~\textit{vs.}~\#3, \#2~\textit{vs.}~\#4, we can find that the ellipse label works better than the regular rectangle labels. According to all these experimental results and analyses, we can conclude that the carefully designed loss functions and classification labels can further improve the final tracking results without using external training samples.

\begin{table}[!htp] 
\caption{Experimental results for the ablation study of Loss Functions and Classification Labels.}
\centering
\begin{tabular}{l|c|c|c|c}
\toprule
Index    & Loss  &Label       & Test AO(\%) & Val AO(\%) \\  
\midrule
1 & GIOU + BCE + L1   & Rectangle  & 71.5   &   84.5       \\
2 & CIOU + BCE + L1   & Rectangle  & 72.0   &  84.8       \\
3 & GIOU + BCE + L1   & Ellipse    & 72.1   &  84.8      \\
4 & CIOU + BCE + L1   & Ellipse    & 72.7   &  85.6      \\
 \bottomrule
\end{tabular} 
\label{ablation4}
\end{table}

\subsection{Parameter Analysis}  
In this section, we will give a detailed introduction to the parameter analysis, including the number of GPHA layers and model parameters.

\noindent
\textbf{Analysis of Different Numbers of GPHA Layers. } 
To further verify the effectiveness of our Spatial-Frequency Transformer, we test our model using different layers of GPHA blocks on the GOT-10K dataset. As shown in Table~\ref{ablation3}, the best tracking results can be obtained when 6 GPHA layers are used, i.e., the AO is 72.7\% on the GOT-10K dataset. When only 2 layers are adopted, the tracking efficiency is the best, i.e., 35 FPS.

\begin{table}[!htp]
\centering 
\caption{Tracking results and speeds with different GPHA layers on GOT-10K benchmark.} 
\begin{tabular}{lllllll}   \toprule
L       &2           & 3       & 4       &5        & 6                & 7 \\  \midrule
AO(\%) &69.3        & 69.8   & 70.9   & 71.1   & \textbf{72.7}   & 72.5     \\
FPS    &\textbf{35}  &32       & 30      & 29      &  27              & 24     \\  \bottomrule
\end{tabular}
\label{ablation3}
\end{table}

\noindent
\textbf{Analysis of Module Parameters. } 
\textcolor{black}{
Table~\ref{got10ktable} reported the parameters of existing trackers, and our proposed SFTransT contains 29.6M parameters. In detail, the cross-scale SwinTransformer, the four MHCA blocks, and the tracking head of SFTransT separately cover 12.6M, 6M, and 1.4M. The proposed SFFormer network covers 9.6M parameters, with each GPHA block containing 1.6M. Therefore, it indicates that our tracker achieves outstanding performance with few parameters and speed costs, validating the effectiveness of our SFTransT and SFFormer. }

\subsection{Attribute Analysis}

In this section, we report the experimental results under challenging factors defined in the LaSOT and TNL2K dataset, including \textit{aspect ration change, adversarial samples, low resolution, partial occlusion, rotation, scale variation,} etc. As shown in Fig.~\ref{lasot_attributes}, our tracker achieves better results even compared to recent SOTA tracking models, such as the MixFormer-1k and TransT, on the LaSOT dataset. The results reported in Fig.~\ref{TNL2K_attributes} also validated the effectiveness of our tracker and beat many SOTA trackers, including SiamR-CNN, SuperDiMP, and AutoMatch.

\begin{figure}[!b]
\centering
\includegraphics[width=1\columnwidth]{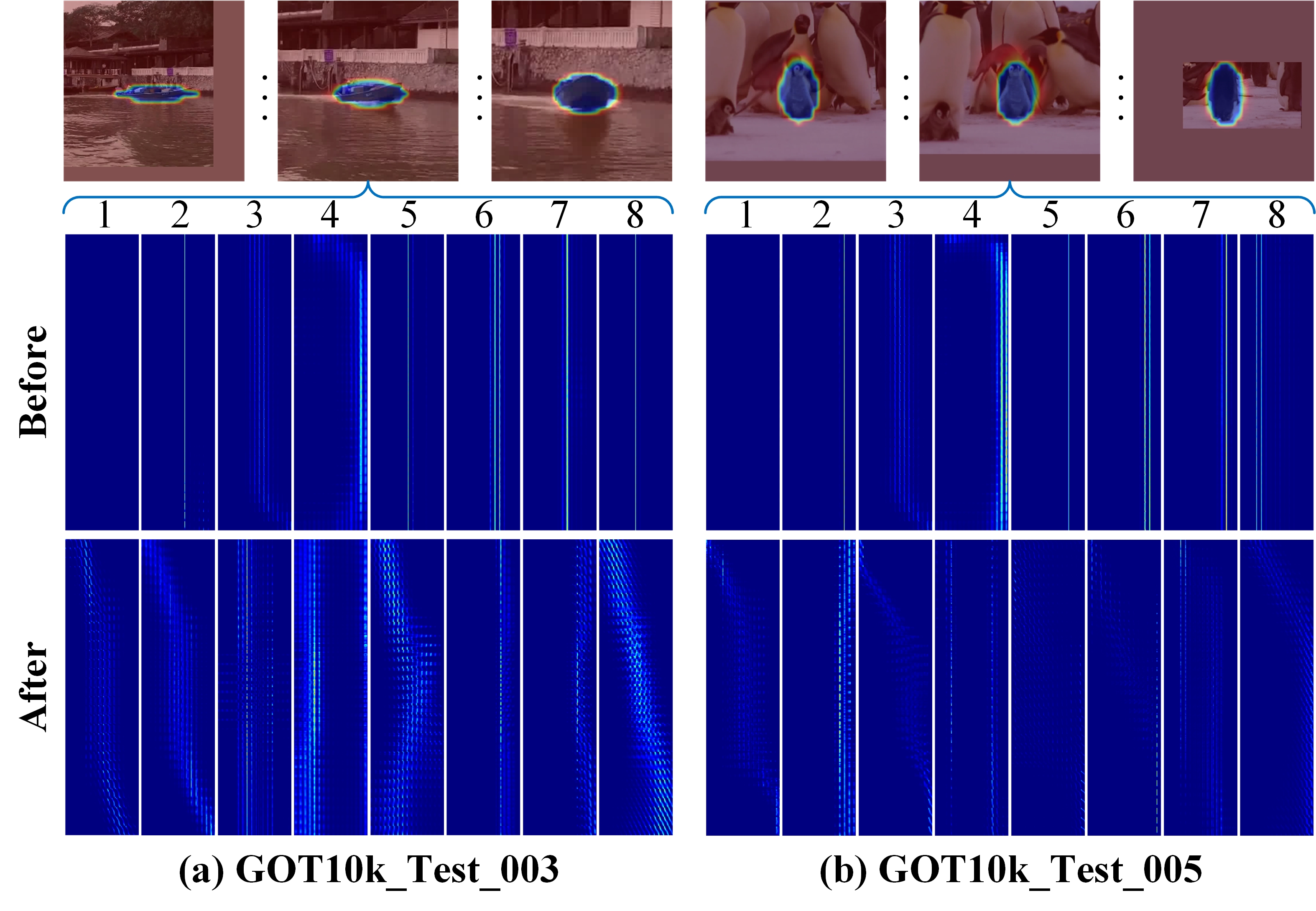} 
\caption{Visualization of the attention weight of multi-head spatial-frequency attention mechanism. Best viewed in color and zoom in.}
\label{attention}
\end{figure}

\subsection{Visualization} 

In addition to the above quantitative analysis, in this subsection, we also give a visualization of spatial-frequency attention and tracking results as the qualitative analysis to help readers better understand our tracker.

\begin{figure*}[!htp]
\centering
\includegraphics[width=1\textwidth]{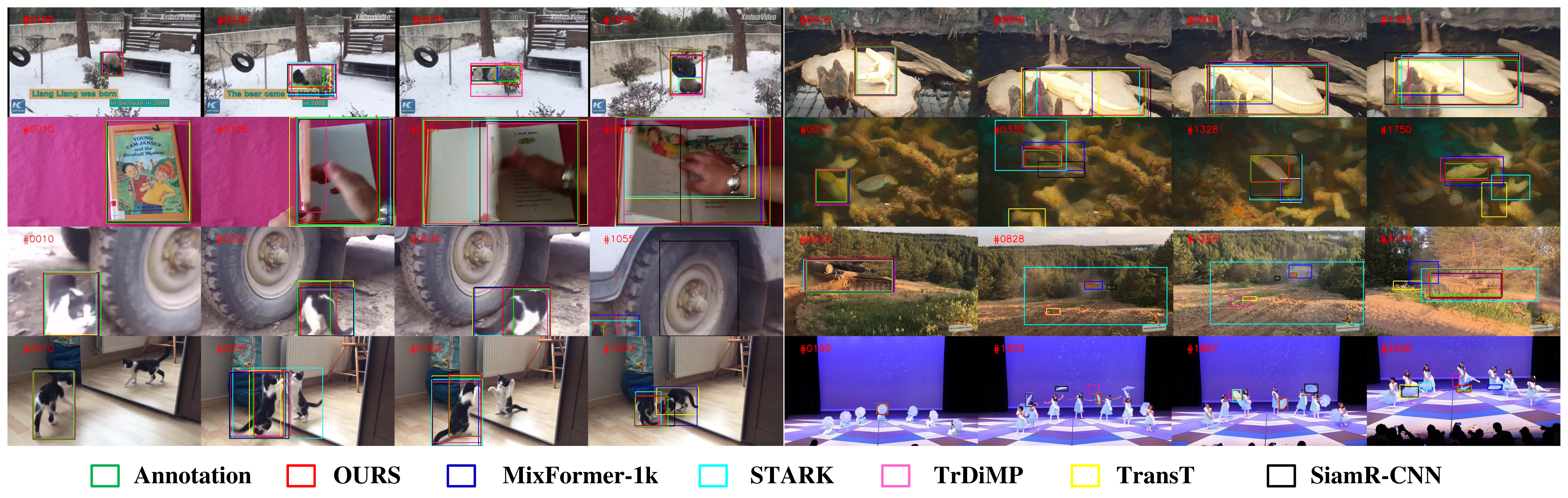} 
\caption{Visualization of tracking results of our proposed and other trackers on LaSOT dataset. Best viewed in color. } 
\label{visualization}
\end{figure*}

\begin{figure}[!htp]
\centering
\includegraphics[width=1\columnwidth]{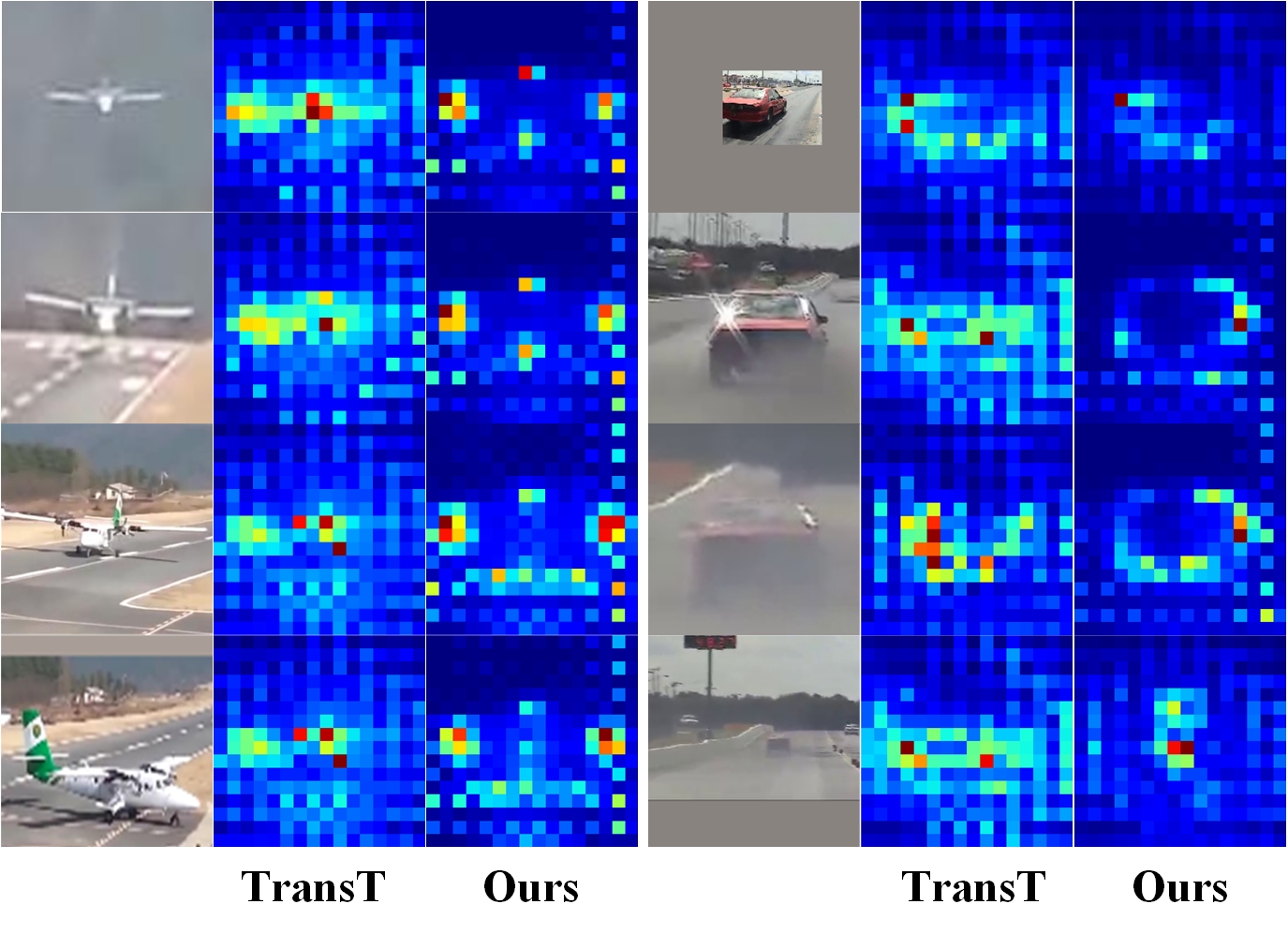} 
\caption{Visualization of activation maps of our SFTransT and TransT. Best viewed in color.}
\label{attn_visualization}
\end{figure}

\noindent
\textbf{Spatial-Frequency Attention. }  
For a more specific analysis of spatial-frequency attention, we give a visualization of head-specific attention weight and classification activation map, as shown in Fig.~\ref{attention}. 
The first row is the video frame with spatial prior. It is easy to find that our proposed SFFormer can focus on the center of search regions. 
The last two rows denote the specific head information representation without and through our SFFormer network. $N$ represents the head index of attention blocks, and the dimension of each subfigure is $S_I \times S_T$. We can easily find that the former row which without our SFFormer contains much more DC components, especially the $2^{th}, 5^{th}, 6^{th}, 7^{th}, 8^{th}$ nearly pure DC bias information. 
Thanks to the proposed SFFormer, which enhances the high-frequency features effectively, we can not only protect the DC bias ($6^{th}$) but also emphasize the high-frequency feature in some heads. These visualized attention maps fully validated that our SFFormer emphasizes high-frequency branch information and effectively avoids the degeneration of target object attention.  

In addition, we also visualize and compare the activation maps with the TransT, as illustrated in Fig.~\ref{attn_visualization}. We can find that our activation maps focus more on the edge position and specific target components. Therefore, our tracker is more accurate than the SOTA tracker TransT.

\noindent
\textbf{Tracking Results. } 
As shown in Fig.~\ref{visualization}, we give a visual comparison of the tracking results of our tracker and five other SOTA trackers, including MixFormer-1k, STARK, TrDiMP, SiamR-CNN, and TransT. It is intuitive to observe that our tracker can be more robust when facing occlusion, deformation, similar distractors, scale variation, fast motion, etc.

\begin{figure}[!htp]
\centering
\includegraphics[width=1\columnwidth]{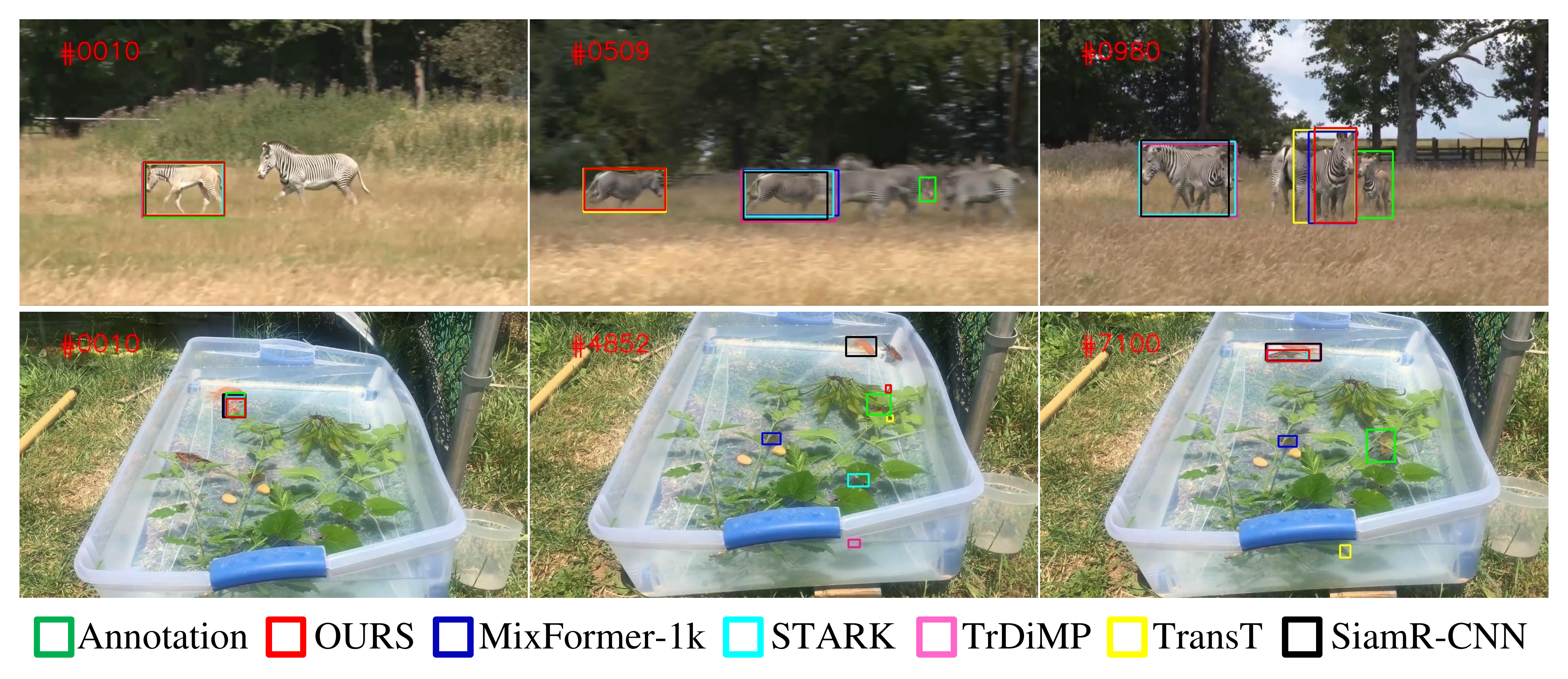} 
\caption{Failed cases of our tracker and other state-of-the-art trackers.}
\label{failure}
\end{figure}

\subsection{Failure Cases and Future Works} 
We believe the analysis above can prove the robustness and accuracy of our proposed spatial-frequency Transformer tracker. At the same time, our tracker hasn't explicitly modelled distractors for dense and small-scale targets. Therefore, it is still easily affected by these factors. We give some visualization of the failed cases in Fig.~\ref{failure}. 
\textcolor{black}{
In our future works, to explore the potential relationship between target and distractors, we will consider adopting spatio-temporal preserving representation~\cite{zhang2019making} and semi-supervised methods~\cite{luo2017adaptive, chen2019semisupervised} to discriminate diverse latent patterns of target and distractors' trajectory. 
}

\section{Conclusion} 
In this paper,  we propose a novel Spatial-Frequency Transformer (SFFormer) network for the visual object tracking task. Precisely, our proposed tracking framework is termed SFTransT, and it contains four main modules, including the cross-scale fusion-based Swin-Transformer, MHCA, SFFormer, and tracking head. The Swin-Transformer is adopted as our backbone network for feature extraction. The MHCA boosts interactive feature learning between the search and template branches. The SFFormer is designed to jointly model Gaussian spatial prior and low-/high- frequency information. The enhanced features will be fed into the tracking head for target object localization. Extensive experiments on multiple benchmark datasets all validated the stable improvement of our proposed tracker. 
\textcolor{black}{
In future works, we will consider designing a Transformer network to integrate spatial, frequency, and temporal representations to explicitly model the target object's appearance with discriminative learning to suppress the influence of distractors and background clutter. 
}


\bibliographystyle{IEEEtran}
\bibliography{track}

\vfill

\end{document}